\documentclass[preprint,10pt]{elsarticle}

\usepackage{amssymb}
\usepackage[utf8]{inputenc}
\usepackage{amsmath}
\usepackage{float}
\usepackage{tikz}
\usepackage{hyperref}
\usepackage[margin=1in]{geometry}
\usepackage{caption}
\usepackage{subcaption}
\usepackage{multirow}

\journal{Physica D: Nonlinear Phenomena}

\begin{document}

\begin{frontmatter}



\title{Learning Governing Equations of Unobserved States in Dynamical Systems}


\author[inst1]{Gevik Grigorian}\corref{cor}

\affiliation[inst1]{organization={Department of Mechanical Engineering},
            addressline={University College London}, 
            city={London},
            postcode={WC1E 6BT},
            country={UK}}
\cortext[cor]{Corresponding author.}
\ead{gevik.grigorian.18@ucl.ac.uk}

\author[inst2,inst3]{Sandip V. George}
\author[inst3]{Simon Arridge}

\affiliation[inst2]{organization={Department of Physics},
            addressline={University of Aberdeen}, 
            city={Aberdeen},
            postcode={AB24 3FX},
            country={UK}}

\affiliation[inst3]{organization={Department of Computer Science},
            addressline={University College London}, 
            city={London},
            postcode={WC1E 6BT},
            country={UK}}

\begin{abstract}
Data-driven modelling and scientific machine learning have been responsible for significant advances in determining suitable models to describe data. Within dynamical systems, neural ordinary differential equations (ODEs), where the system equations are set to be governed by a neural network, have become a popular tool for this challenge in recent years. However, less emphasis has been placed on systems that are only partially-observed. In this work, we employ a hybrid neural ODE structure, where the system equations are governed by a combination of a neural network and domain-specific knowledge, together with symbolic regression (SR), to learn governing equations of partially-observed dynamical systems. We test this approach on two case studies: A 3-dimensional model of the Lotka-Volterra system and a 5-dimensional model of the Lorenz system. We demonstrate that the method is capable of successfully learning the true underlying governing equations of unobserved states within these systems, with robustness to measurement noise.
\end{abstract}


\begin{keyword}
Scientific machine learning \sep Hybrid neural ODE \sep Dynamical systems \sep Symbolic regression \sep Time series
\end{keyword}

\end{frontmatter}


\section{Introduction}
\label{Introduction}


Models based on dynamical systems are useful in describing time-dependent phenomena in the real world. Continuous time processes are described using dynamical models based on differential equations. Discovering the underlying governing equations from observed time series data is a problem that has generated considerable interest, particularly with the recent thrust towards data-driven modelling. Such approaches use input from data, instead of {\it a priori} knowledge, to determine a useful model for the system. An alternate approach is to use data as well as system knowledge to determine appropriate ordinary differential equation (ODE) models. A specific area of interest in the context of the latter is to determine unknown terms of a system of ODEs using observed time series data of the system. Such questions appear in various different areas of application including biology\cite{prokop2024biological}, chemistry\cite{sorourifar2023physics, lima2023improved}, engineering\cite{mackay2023informed} and the geosciences\cite{shen2023differentiable}

A number of approaches have been developed recently in scientific machine learning to determine unknown terms in a system of partially-known differential equations\cite{brunton2016discovering}. The framework of universal differential equations (UDEs) is one such approach, where unknown terms in a system of differential equations are replaced by neural networks\cite{rackauckas2020universal}. These neural networks can be trained on time series data derived from the system, to fully capture 
its dynamics.
This approach is often referred to as a hybrid neural ODE. It is also possible to determine the terms missing from the system through regression of the trained neural network, using techniques such as the sparse identification of nonlinear dynamics (SINDy) or symbolic regression (SR) \cite{brunton2016discovering, bongard2007automated}.

While considerable focus has been placed on using hybrid neural ODE approaches to determine unknown terms in partially-known but fully observed systems of differential equations, a more challenging problem is to study systems of differential equations that are only partially-observed. A possible example of this could be the population dynamics of a system of interacting species, where the population of one is difficult to measure. In accordance with Taken's embedding theorem, information about the dynamics of the unobserved variable must be fully captured in the dynamics of the other variables\cite{takens2006detecting}. 

In the recent past, there has been a focus on addressing this question. In \cite{somacal2022uncovering}, the authors attempted to discover the terms involved in an equation corresponding to a single variable when only time series data from that variable was available, using higher-order time derivatives and Lasso regression. The latter is a regression technique that accounts for model simplicity by promoting sparse solutions. In \cite{bakarji2023discovering} autoencoding and the sparse identification of nonlinear dynamics (SINDy) algorithm were used to discover alternate equations for the Lorenz system when only one of the variables of the system was observed.

In this paper, we use the framework of hybrid neural ODEs and SR to tackle the problem of partially-observed systems. We start by simulating time series data from the 3-dimensional Lotka-Volterra and 5-dimensional Lorenz systems.

The systems are chosen to demonstrate the applicability of the method over a differing number of variables and dynamical behaviours. Next, at least one of the variables is considered to be unobserved. The equation(s) for the unobserved variables are also assumed to be unknown and hence replaced by neural networks, resulting in a hybrid neural ODE. The hybrid neural ODE is then trained using the observed variables, using the mean squared error of the predicted values from the observed values as the loss function. Finally, SR is employed to determine the unknown equations in the model.

\section{Methodology} \label{Methodology}

The Lotka-Volterra equations, defined in section \ref{3DLotkaVolterra}, are a system of nonlinear coupled ODEs which describe predator-prey interactions, a key phenomenon in ecology. The Lorenz equations, defined in section \ref{5DLorenz}, are a system of nonlinear coupled ODEs which exhibit chaotic behaviour and are used to model thermal convection. The details of both the training process of the Lotka-Volterra/Lorenz hybrid neural ODEs and the SR stage are presented in section \ref{TrainingRegression}. Training was carried out in Julia \cite{bezanson2017julia}, where the original systems and the hybrid neural ODEs were simulated using the Tsit5 ODE solver, with both relative and absolute tolerances of $10^{-6}$. SR was implemented using the Python package (with Julia back-end) PySR \cite{cranmer2023interpretable}. Experiments were conducted both without and in the presence of measurement noise. In this work, Gaussian noise at various percentages of the standard deviation of the synthetic data were added to the data.

\subsection{3D Lotka-Volterra System} \label{3DLotkaVolterra}

\begin{figure}[H]
    \centering
    \includegraphics[scale=0.5]{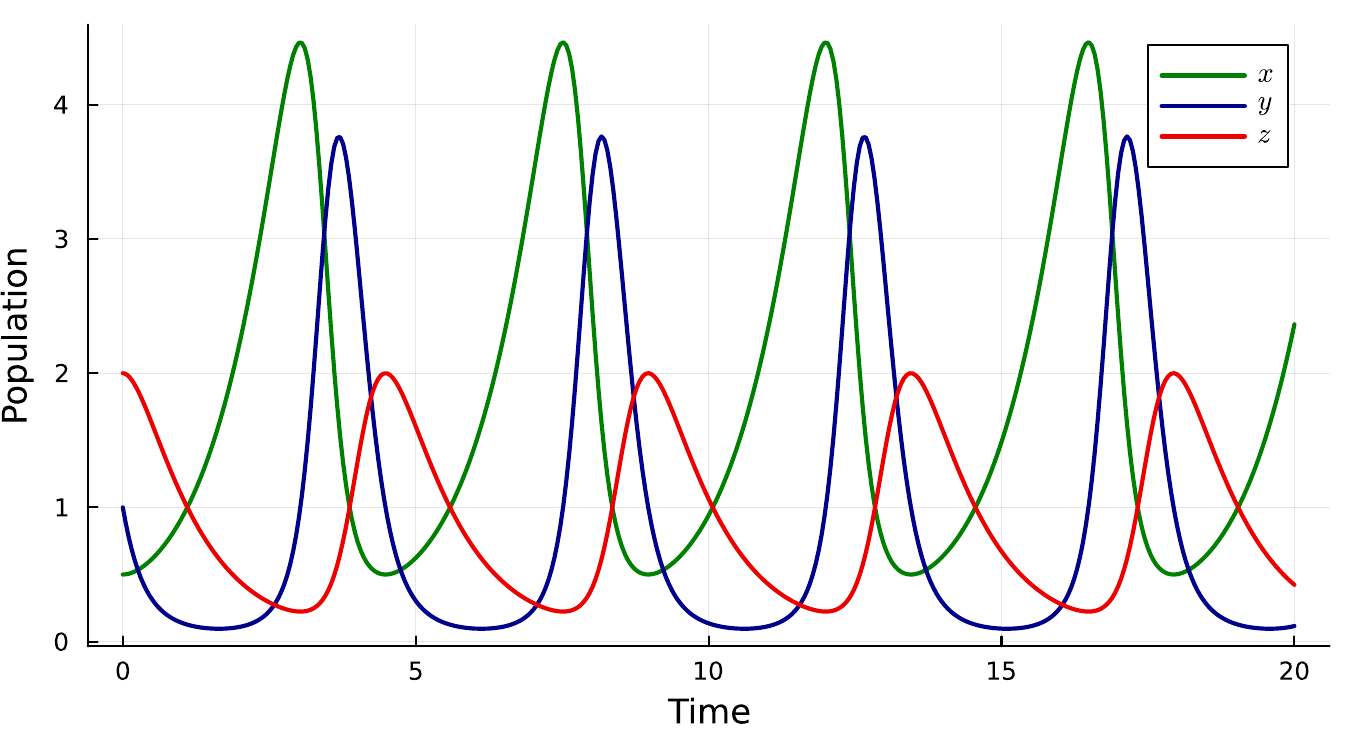}
    \caption{\label{fig:LV3} Temporal evolution of the 3D Lotka-Volterra system with initial condition $[x_0,y_0,z_0]=[0.5,1.0,2.0]$. This synthetic data is used for training and validation.}
\end{figure}

Many variants of the Lotka-Volterra equations have been proposed, but we employ a simple 3-dimensional version, describing the interactions between 3 species in a system. The model equations are given by

\begin{align} 
    \frac{dx}{dt} &= ax - bxy, \tag{1.1} \label{1.1} \\
    \frac{dy}{dt} &= -cy + dxy - eyz, \tag{1.2} \label{1.2}\\
    \frac{dz}{dt} &= -fz + gyz, \tag{1.3} \label{1.3}
\end{align}
\noindent
where $x, y$ and $z$ are the 3 state variables representing the 3 species. The parameters $a, c$ and $f$ represent the growth rates of species $x, y$ and $z$, respectively, and the parameters $b, d, e$ and $g$ represent the effects of predation or competition on their growth rates.
In this work, we use a value of 1.0 for all parameters $a, b, c, d, e, f, g$, as well as an initial condition of $x_0 = 0.5, y_0 = 1.0, z = 2.0$. The dynamics of this system are periodic and therefore do not depend significantly on parameter values and initial conditions. This system is simulated for 20 units of time, at a sampling rate of 0.05 (frequency of 20 Hz), generating a synthetic data set of 1200 data points ($400\times3$). Given the periodic behaviour of the system, 20 units of time was selected to capture a number of full periods. The dynamics of this simulation are shown in Figure \ref{fig:LV3}.

We now assume that equation \ref{1.2} is unknown and that measurement data for $y$ are not available. We therefore construct a hybrid neural ODE by replacing equation \ref{1.2} with $\frac{dy}{dt}=N_\theta (x,y,z)$, where $N_{\theta}$ is a neural network parameterised by the set of weights and biases $\theta$. The architecture of the network is a fully connected network with 3 inputs ($x,y, z$), 2 hidden layers of 40 neurons each with Gaussian error linear unit (gelu) activation functions, and a single output. The activation gelu is defined as $\text{gelu}(x)=x\Phi (x),$ where $\Phi(x)$ is the standard Gaussian cumulative distribution function. It can also be approximated by $\text{gelu}(x) \approx 0.5x\left(1 + \text{tanh} \left[\sqrt{\frac{2}{\pi}}\left(x+0.044715x^3\right)\right]\right)$, which is the form used in this work. The single output of the network represents the dynamics of $\frac{dy}{dt}$. The synthetic data is considered as ground truth and a window of this data is selected for training. Given that this system generates periodic dynamics, a training range of less than one period of the data is selected, namely $0-2$ units of time. This means the training set consists of 80 data points (40 points in $x$ and $z$ each, while $y$ is unobserved). The experiment is repeated at 0\%, 2\% and 5\% noise added to the training data.

\subsection{5D Lorenz System} \label{5DLorenz}

\begin{figure}[H]
    \centering
    \includegraphics[scale=0.5]{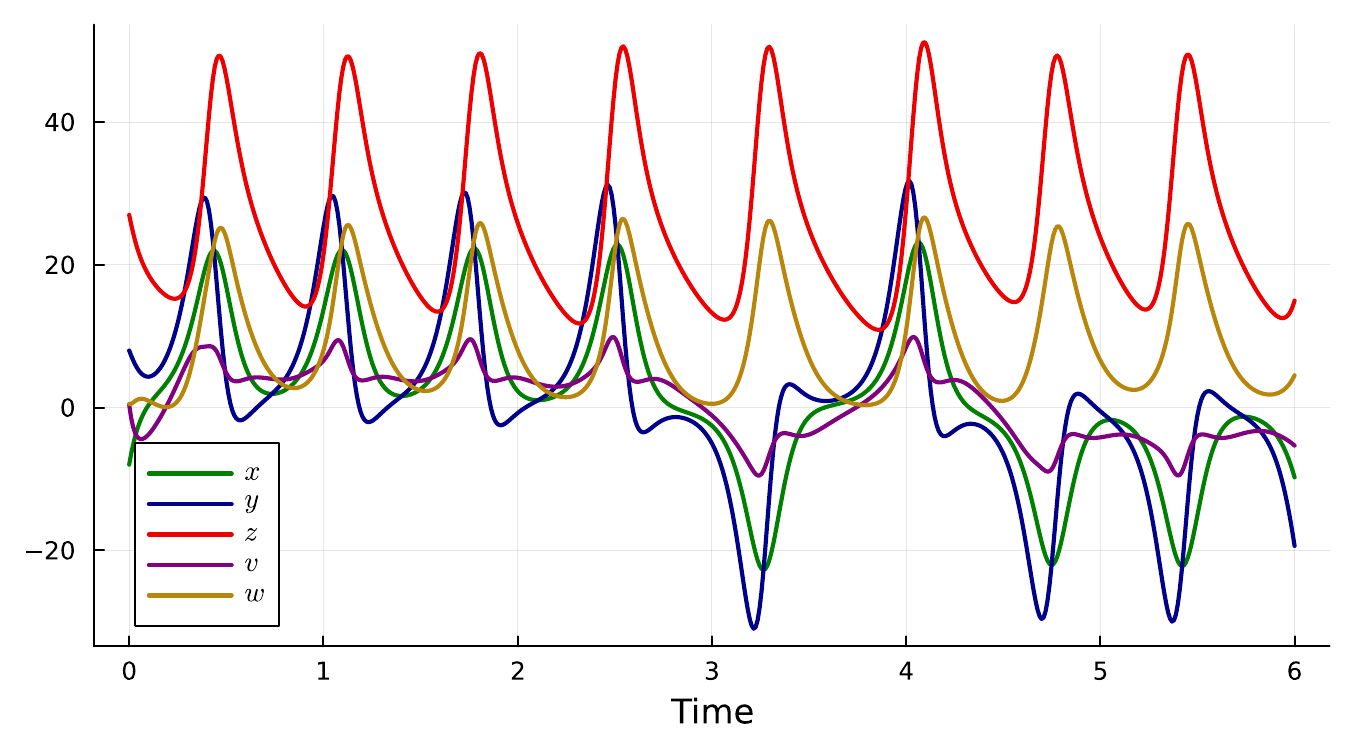}
    \caption{\label{fig:Lorenz5D} Temporal evolution of the 5D Lorenz system with initial condition $[x_0,y_0,z_0,v_0,w_0]=[-8.0,8.0,27.0,0.4,0.5]$. This synthetic data is used for training and validation.}
\end{figure}

We employ the 5-dimensional version of the Lorenz system, defined as

\begin{align}
    \frac{dx}{dt} &= \sigma(y - x), \tag{1.4} \label{1.4}\\
    \frac{dy}{dt} &= x(\rho - z) - y, \tag{1.5} \label{1.5}\\
    \frac{dz}{dt} &= xy - \beta z - xv, \tag{1.6} \label{1.6}\\
    \frac{dv}{dt} &= xz - 2xw - (1 + 2\beta)v, \tag{1.7} \label{1.7}\\
    \frac{dw}{dt} &= 2xv - 4\beta w \tag{1.8} \label{1.8},
\end{align}
\noindent
where the state variables $x,y,z,v,w$  collectively represent the spatial distribution and dynamics of temperature deviations within a convective cell, including horizontal and vertical variations, as well as additional aspects of the system's behavior. $\sigma$ is the Prandtl number, $\rho$ is the Rayleigh number and $\beta$ is the ratio of the width to the height of the convective cell. We use the parameter values $\sigma = 10, \rho = 35, \beta = \frac{8}{3}$ and the initial conditions $x_0 = -8, y_0 = 8, z_0 = 27, v_0 = 0.4, w_0 = 0.5$. These values are common choices in the literature and are selected here as they correspond to a state of the system that lies close to the attractor, minimising any initial transients. A simulation of 6 units of time is carried out, using a sampling rate of 0.01 (frequency of 100 Hz), resulting in a data set of 3000 data points ($600 \times 5$). Since the Lorenz system exhibits chaotic behaviour, 6 units of time is selected in order to capture a significant portion of the dynamics while also concentrating on the range where predictions may deviate from the ground truth. The dynamics of this system are shown in Figure \ref{fig:Lorenz5D}.

We now assume that equations \ref{1.4} and \ref{1.5} are unknown and that measurement data for $x$ and $y$ are unknown. We construct a hybrid neural ODE by replacing equations \ref{1.4} and \ref{1.5} with $\frac{dx}{dt}=N_{\theta}^1(x,y,z,v,w)$ and $\frac{dy}{dt}=N_{\theta}^2(x,y,z,v,w)$, respectively. Here, $N_\theta$ is a network parameterised by the set of weights and biases $\theta$. The architecture of the network is a fully connected network with 5 inputs ($x,y,z,v,w$), 2 hidden layers of 40 neurons with gelu activation functions (defined in section \ref{3DLotkaVolterra}), and 2 outputs. The two outputs are specified by the superscripts 1 and 2. Again, the synthetic data is considered as ground truth and the training range was selected to be the data between 0 - 0.25 units of time, resulting in a training data set of 75 points (25 points in $z, v$ and $w$ each, with $x$ and $y$ being unobserved). The chaotic nature of this system together with the task of learning two governing equations, poses a significantly more challenging problem than that which is outlined in section \ref{3DLotkaVolterra}. Given this added difficulty, this experiment is only repeated at 0\% and 0.3\% noise added to the training data.

\subsection{Methodology Details} \label{TrainingRegression}

For the problems outlined in sections \ref{3DLotkaVolterra} and \ref{5DLorenz}, the training process, the implementation of SR and further analysis on the lengths of the training ranges are detailed in this section.

\subsubsection{Training} \label{Training}

The hybrid neural ODEs have the form
\begin{equation}
\dot{\bf U} = \begin{pmatrix}F_{ {\bf C}}\\N_{\theta}\end{pmatrix}({\bf U}),
\end{equation}

where ${\bf U}$ is the vector of state variables ($[x,y,z]$ for the Lotka-Volterra case and $[x,y,z,v,w]$ for the Lorenz case), $F_C$ is a set of prescribed mathematical operations with parameters $C$ (equations \ref{1.1} and \ref{1.3} for the Lotka-Volterra case, and equations \ref{1.6}, \ref{1.7} and \ref{1.8} for the Lorenz case) and $N_\theta$ is a neural network, parameterised by $\theta$. A normal ODE corresponds to the case where the vector of the states is related to its derivative through mathematical operations ($F_C$) only. A neural ODE corresponds to the case where the state vector is relative to its derivative through a neural network ($N_\theta$) only. Hence, in this case, a hybrid neural ODE refers to the relationship between the state vector and its derivative being governed by some combination of mathematical operations and a neural network.

The training process involves simulating the hybrid neural ODE for the length of the training range (using the same solver, tolerances, and sampling rate) in order to generate a prediction. During training, the unobserved states are removed from both the ground truth data and the prediction. The remaining values (only the observed states) from the ground truth and the prediction are then compared in order to generate a loss value. The loss function used in this work is mean squared error (MSE), defined as

\begin{equation}
    L = \frac{1}{D} \sum_{i=1}^{D} (u_i - \hat{u}_i)^2,
\end{equation}
\noindent
where $L$ is the loss value, $D$ is the number of data points, $u_i$ is the $i$th element of the ground truth (after removing the unobserved states) and $\hat{u}_i$ is $i$th element of the prediction (after removing the unobserved states).

Training is then the process of minimising this loss function by optimising the weights and biases of the network. The optimisation is done in two stages, as in \cite{rackauckas2020universal}. First, Adam \cite{kingma2014adam} is used to move the network parameters into a more favourable starting position for the L-BFGS optimiser \cite{liu1989limited}, which is used subsequently. Adam is a common choice as it is easy to implement and works efficiently with little hyper-parameter tuning. At this stage, the Adam optimiser is used with a learning rate of 0.01 for 1000 iterations. However, Adam may sometimes converge slowly and therefore the L-BFGS optimiser (a quasi-Newton algorithm) is used subsequently for faster convergence (and sometimes towards a better optimum), as it uses second-order information about the loss function (the Hessian matrix) to make more informed updates
to the network parameters. For this second stage, the L-BFGS optimiser is used for 1000 iterations.

In this work, an ensemble of ten hybrid neural ODEs is trained. This means that for each of the cases outlined in sections \ref{3DLotkaVolterra} and \ref{5DLorenz}, the training process is carried out 10 times. Due to the random initialisation of the network parameters, the training process results in a different set of learned parameters each time and therefore different predictions. Averaging over these predictions is a means of improving accuracy and obtaining desired results more consistently. In this work, a mean of the outputs of the ten networks in the ensemble is taken. This training process is repeated in the same way at each level of noise.

\subsubsection{Symbolic regression} \label{SymbolicRegression}

Upon completing the training process, any of the ten hybrid neural ODEs in the ensemble can be used to make extrapolations (predictions beyond the range of training data). While these predictions are expected to be more accurate than that of a black box model (due to the physical knowledge encoded in the system equations), they can still lack some predictive accuracy since neural networks are known to be poor at extrapolating. This would especially be true if the training data captures only a small portion of the system dynamics. This motivates the use of a sparse regression technique, where the neural network in the hybrid neural ODE is converted to symbolic expressions. This step results in a partially-learned model with greater interpretability than the hybrid neural ODE, but can also improve the extrapolation capabilities.

In this work, SR is used for this step. SR is a machine learning technique that fits analytic expressions to data. It requires a set of unary operators (e.g. sin, cos, exp, etc) and a set of binary operators (e.g addition, subtraction, multiplication, division, etc) to be defined by the user. The function space defined by these operators is then searched in a ‘brute force’ manner via genetic programming. Processes such as mutations, crossovers and tournaments encourage a ‘survival of the fittest’ environment among different candidate expressions. For a more detailed description of SR, see \cite{augusto2000symbolic} and \cite{wang2019symbolic}.

SR requires a set of inputs and a target set (the data to fit an expression to). First, each of the 10 hybrid neural ODEs in the ensemble are simulated for the length of the training range. The mean of the inputs to the networks ($[x,y,z]$ for the Lotka-Volterra problem and $[x,y,z,v,w]$ for the Lorenz problem) is taken. This averaged input is treated as the input to SR. The mean of the outputs of the networks ($[x,y,z]$ for the Lotka-Volterra problem and $[x,y,z,v,w]$ for the Lorenz problem) is also taken. This averaged output is treated as the target data for SR. Table \ref{tab:PySRdetails} in Appendix \ref{Appendix2} shows the hyper-parameters used in the PySR implementation.

\subsubsection{Varying the lengths of the training range}
For both cases, further analysis is also done to investigate the effects of varying the length of the training range on the accuracy of the extrapolations. To this end, a hybrid neural ODE is trained using three different training ranges for each case study. For the Lotka-Volterra case, the training ranges are 2.0, 3.25 and 4.5 units of time, while for the Lorenz case, the training ranges are 0.4, 0.8 and 1.2 units of time. 
For each training range (and each level of noise), a sliding window approach is used, where the root MSE (RMSE) between the prediction and the ground truth is calculated within a window of 20 data points. The errors accumulate as the window shifts along to the end of the simulation, and these errors are compared.

\section{Results} \label{Results}

Once each of the ten hybrid neural ODE models in the ensemble are trained, the untrained and trained predictions of a randomly selected model from the ensemble is shown, as well as an extrapolation to examine the model's ability to generalise beyond the training range. Corresponding extrapolations of the partially-learned models trained on noisy data are given in Appendix \ref{Appendix}.

\subsection{3D Lotka-Volterra System\label{sec:results_LV}}

\begin{figure}[H]
    \centering
\begin{subfigure}{0.49\linewidth}
\centering
    \includegraphics[width=\linewidth]{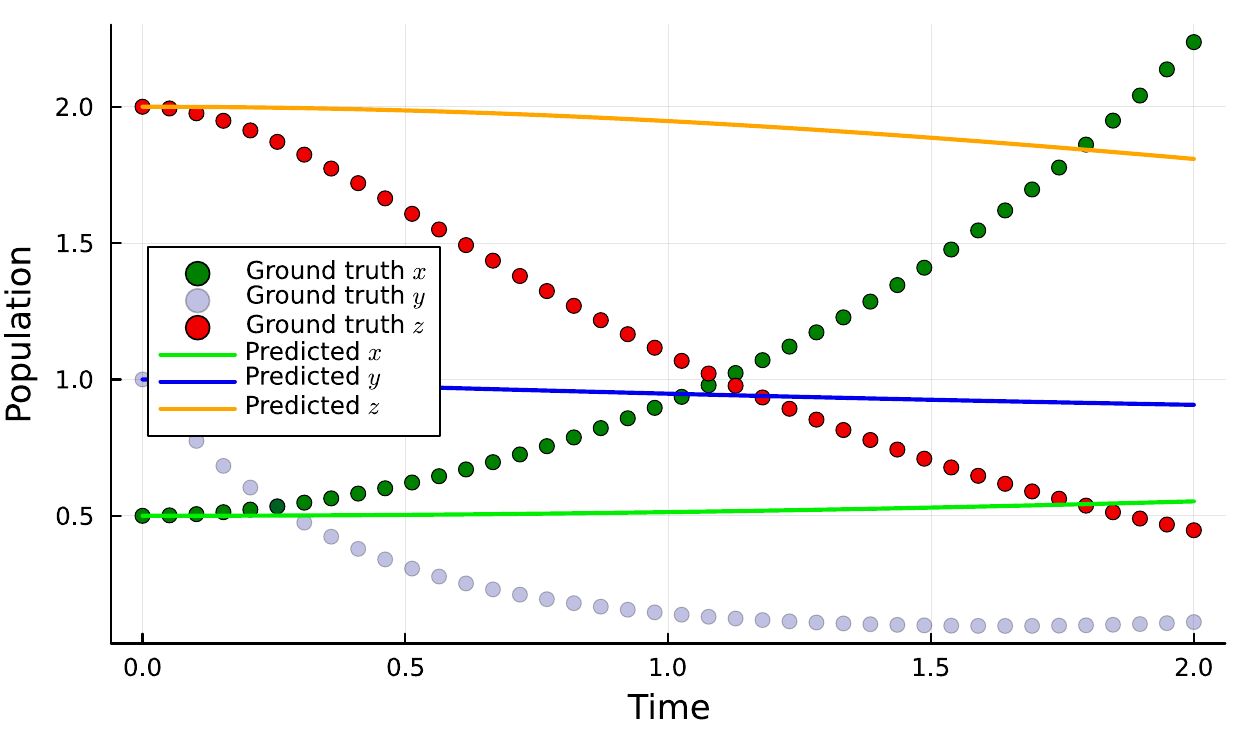}
    \caption{Untrained hybrid neural ODE prediction}
    \label{fig:LV3Untrained}
\end{subfigure}
\begin{subfigure}{0.49\linewidth}
\centering
    \includegraphics[width=\linewidth]{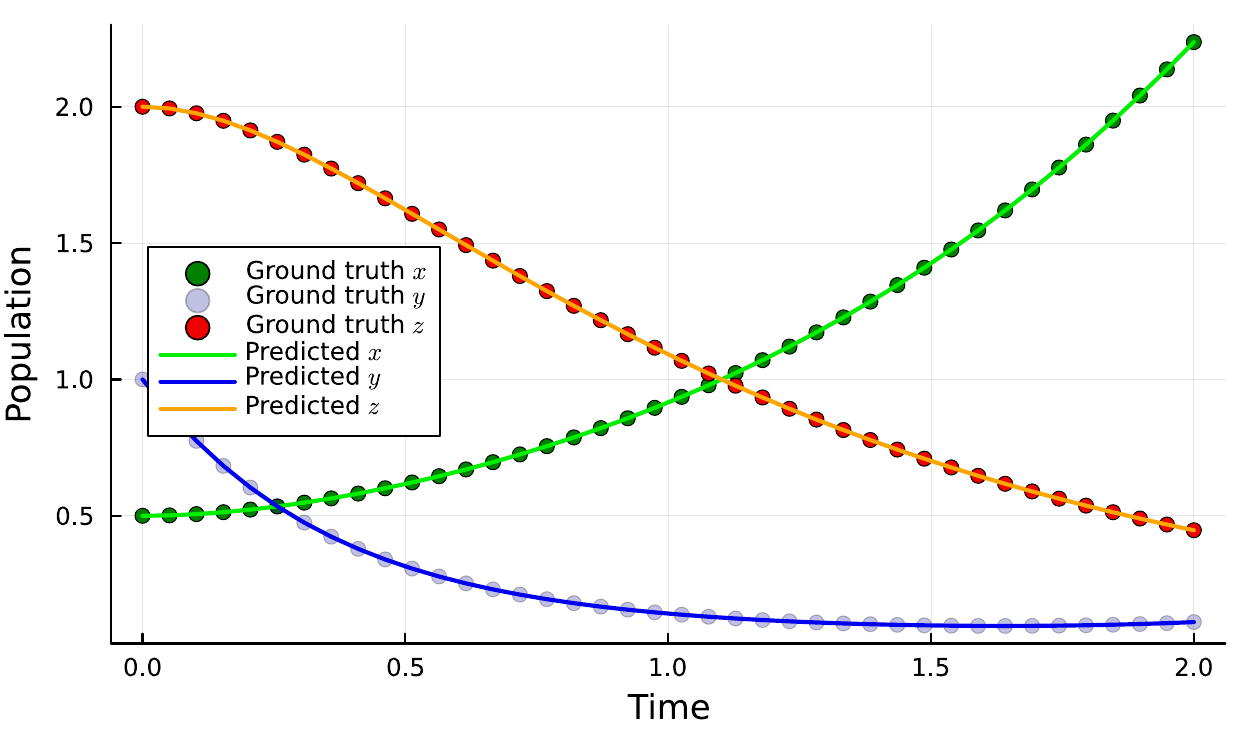}
    \caption{Trained hybrid neural ODE prediction}
    \label{fig:LV3Trained}
\end{subfigure}
    \caption{Prediction of the hybrid neural ODE (shown by the solid curves) against the ground truth data (shown by the scatter points), both before (Figure \ref{fig:LV3Untrained}) and after (Figure \ref{fig:LV3Trained}) the training process. The ground truth data for the state variable $y$ is faint as this data is not available during training.}
    \label{fig:LV3Training}
\end{figure}

Figures \ref{fig:LV3Untrained} and \ref{fig:LV3Trained} show the predictions of a hybrid neural ODE from the ensemble before and after the training process, respectively. In both of these plots, the ground truth of the $y$ state variable is faint to emphasise that this data is not available to the model during training.

\begin{figure}[H]
    \centering
    \includegraphics[width=0.8\linewidth]{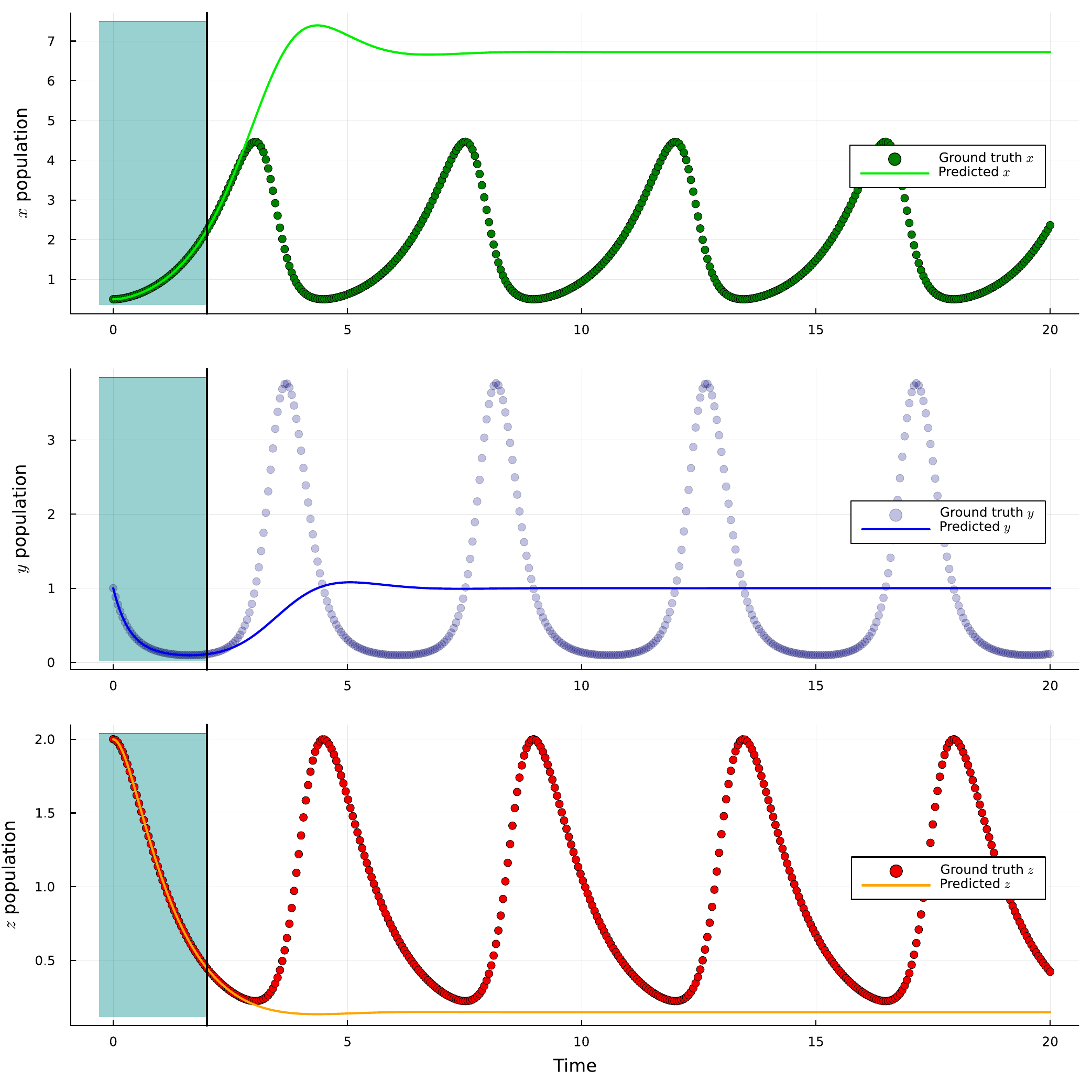}
    \caption{\label{fig:LV3TrainedEx} Extrapolation of trained hybrid neural ODE. The predictions are shown by the solid curves and the ground truth is shown by the scatter points. The ground truth for the state variable $y$ is faint as this data is assumed to be unavailable. The blue region to the left of each plot represents the training range.}
\end{figure}

Figure \ref{fig:LV3Training} shows that the predictions after the training process are very accurate, which is expected given that neural networks are universal function approximators and are therefore powerful at interpolating (making predictions within the training range). An extrapolation of this hybrid neural ODE is subsequently carried out up to 20 units of time, as shown in Figure \ref{fig:LV3TrainedEx}, to examine its ability to generalise beyond the range of training data.

\begin{figure}[H]
    \centering
    \begin{subfigure}{0.49\linewidth}
    \centering
    \includegraphics[width=\linewidth]{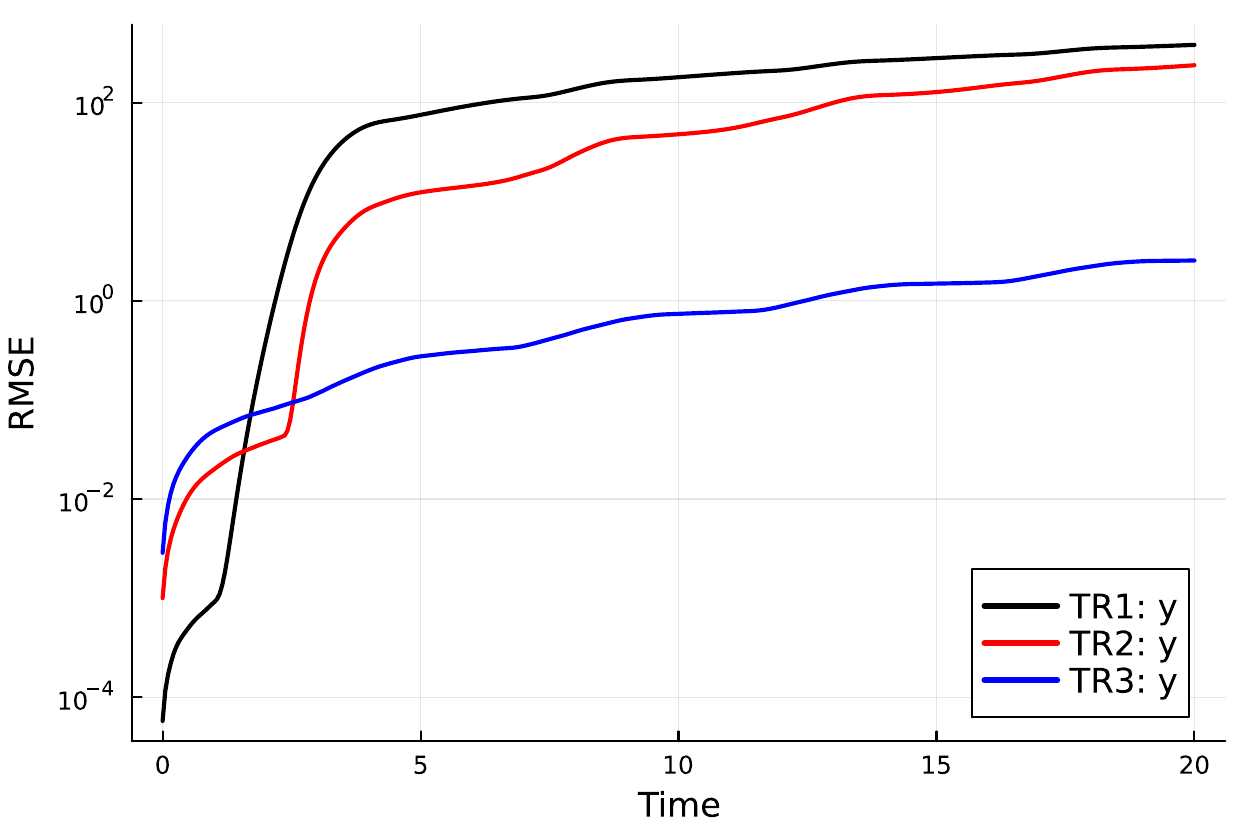}
    \caption{0\% noise}
    \label{fig:LV3Errors0noise}
    \end{subfigure}
    \begin{subfigure}{0.49\linewidth}
    \centering
    \includegraphics[width=\linewidth]{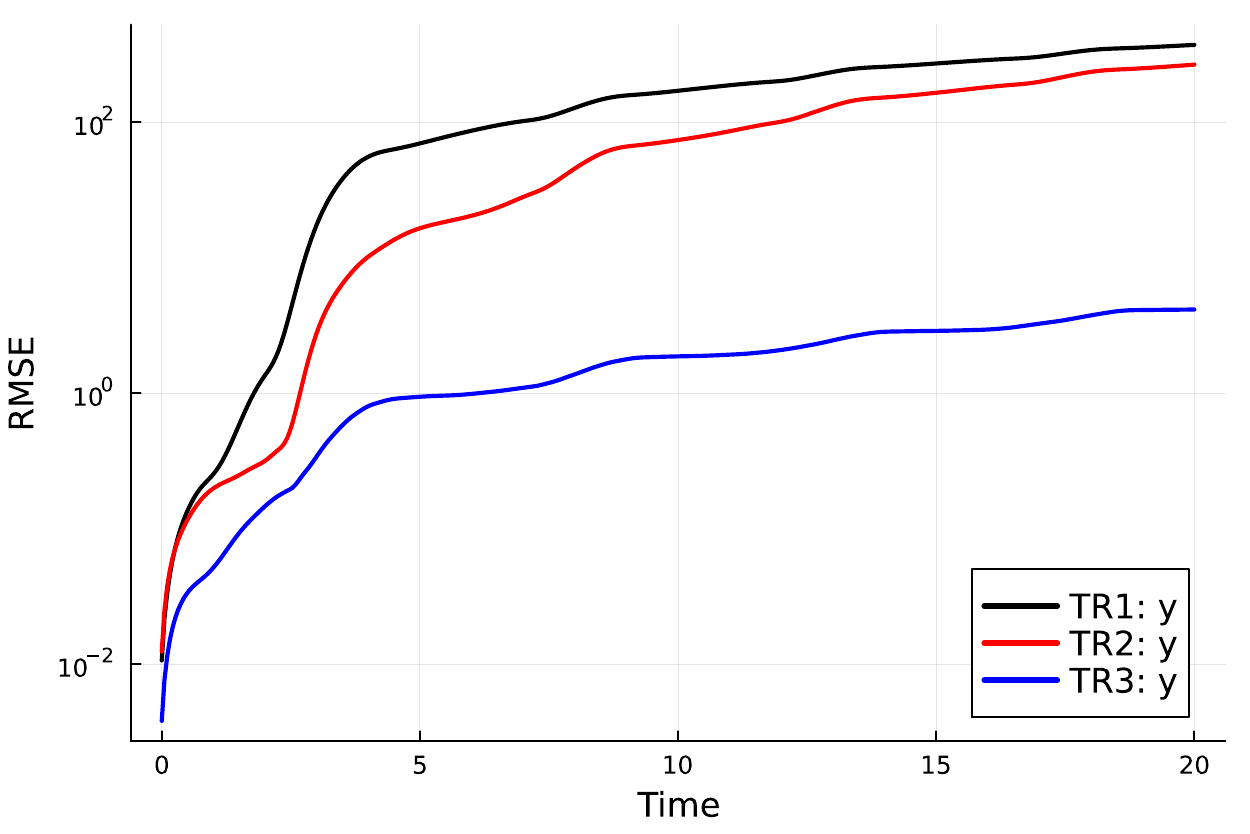}
    \caption{2\% noise}
    \label{fig:LV3Errors2noise}
    \end{subfigure}
    \begin{subfigure}{0.49\linewidth}
    \centering
    \includegraphics[width=\linewidth]{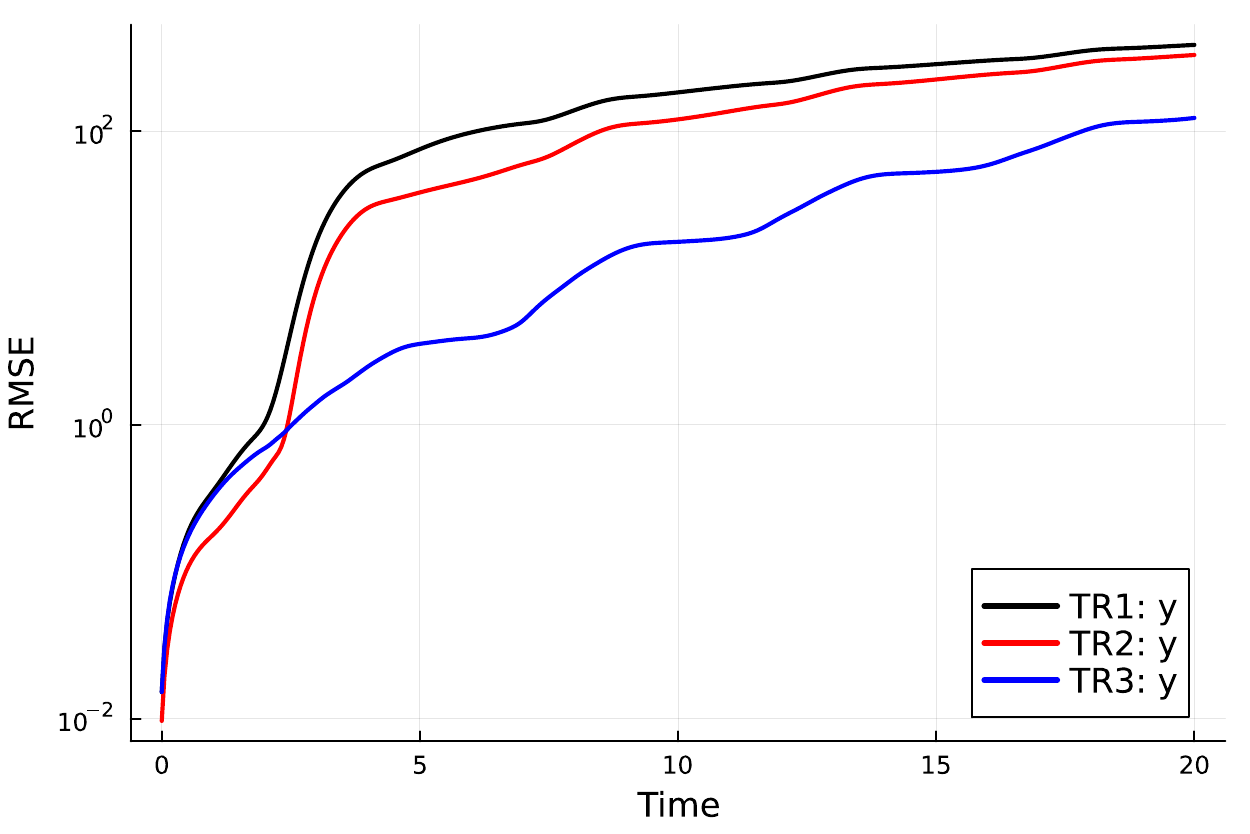}
    \caption{5\% noise}
    \label{fig:LV3Errors5noise}
    \end{subfigure}
    \caption{Sliding window errors of hybrid neural ODE extrapolations, using three different training ranges. The training ranges of 2.0, 3.25 and 4.5 units of time are abbreviated to TR1, TR2 and TR3, respectively. Only the unobserved state variable $y$ is shown. This comparison is done at each level of noise.}
    \label{fig:LV3Errors}
\end{figure}

From Figure \ref{fig:LV3TrainedEx}, it can be seen that the extrapolation of the hybrid neural ODE 
is poor. While neural networks are generally weak at extrapolating beyond their training range, it is expected that the
added domain knowledge present in the hybrid neural ODE (equations \ref{1.1} and \ref{1.3}) may guide the neural network to 
produce more accurate extrapolations. 
However, in this case, the poor extrapolation capabilities of the neural network prevails due to the short range of dynamics captured within the training data. Figure \ref{fig:LV3Errors} shows the errors accumulated via the sliding window approach on the extrapolations of the hybrid neural ODE using three different training ranges. The accuracy of the predictions increases as the training range increases. This trend is expected as a longer training range provides the neural network with more information about the system. The periodic nature of the dynamics also explains this behaviour, since the greater the proportion of a full period of the data that is captured within the training range, the more likely it is to achieve accurate extrapolations. In this work, the longest training range (4.5 units of time) corresponds to roughly one full cycle of the data.

\begin{figure}[H]
    \centering
\begin{subfigure}{0.49\linewidth}
\centering
    \includegraphics[width=\linewidth]{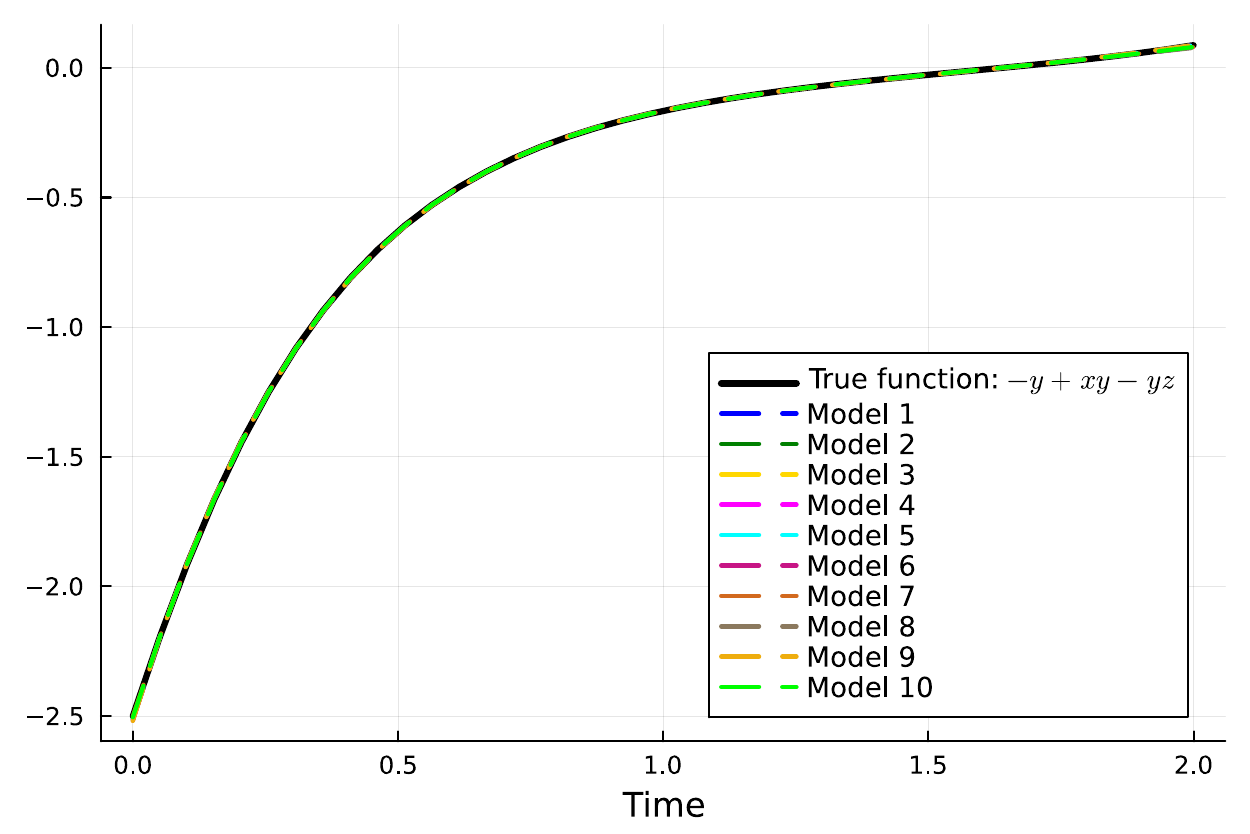}
    \caption{0\% noise}
    \label{fig:LV3Dynamics0}
\end{subfigure}
\begin{subfigure}{0.49\linewidth}
\centering
    \includegraphics[width=\linewidth]{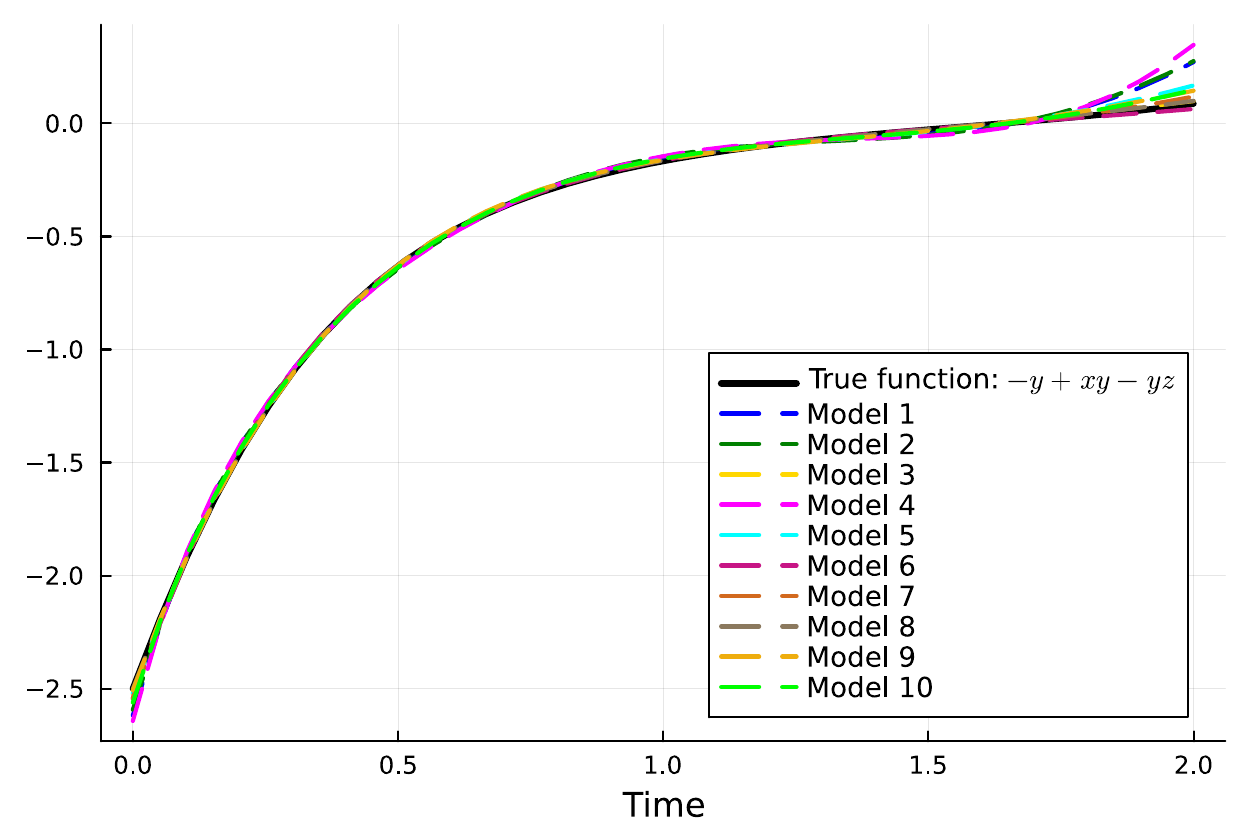}
    \caption{2\% noise}
    \label{fig:LV3Dynamics2}
\end{subfigure}
\begin{subfigure}{0.49\linewidth}
\centering
    \includegraphics[width=\linewidth]{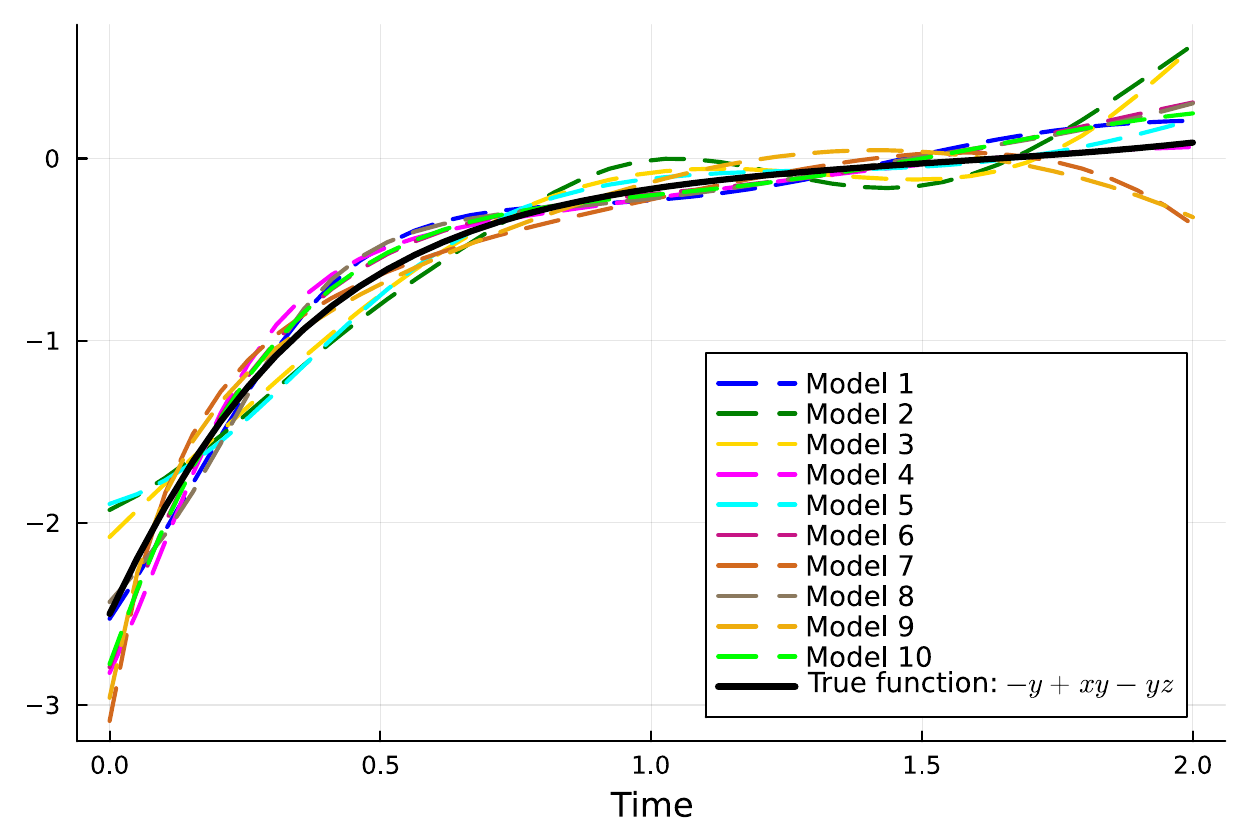}
    \caption{5\% noise}
    \label{fig:LV3Dynamics5}
\end{subfigure}
    \caption{The predicted dynamics of the $\frac{dy}{dt}$ equation generated by each network in the ensemble, at each level of noise added to the ground truth data.}
    \label{fig:LV3Dynamics}
\end{figure}

Returning to the case of a training range of 2.0 units of time, the use of SR is motivated, which is applied to the averaged output of the ten neural networks in the ensemble. To this end, each of the trained hybrid neural ODEs are simulated for the length of the training range to generate the temporal evolution of the states [$x,y, z$], which the neural networks take as input to generate an output representing the learned dynamics for the $\frac{dy}{dt}$ equation. This is done at each level of noise, as shown in Figure \ref{fig:LV3Dynamics}.

\begin{figure}[H]
    \centering
\begin{subfigure}{0.49\linewidth}
\centering
    \includegraphics[width=\linewidth]{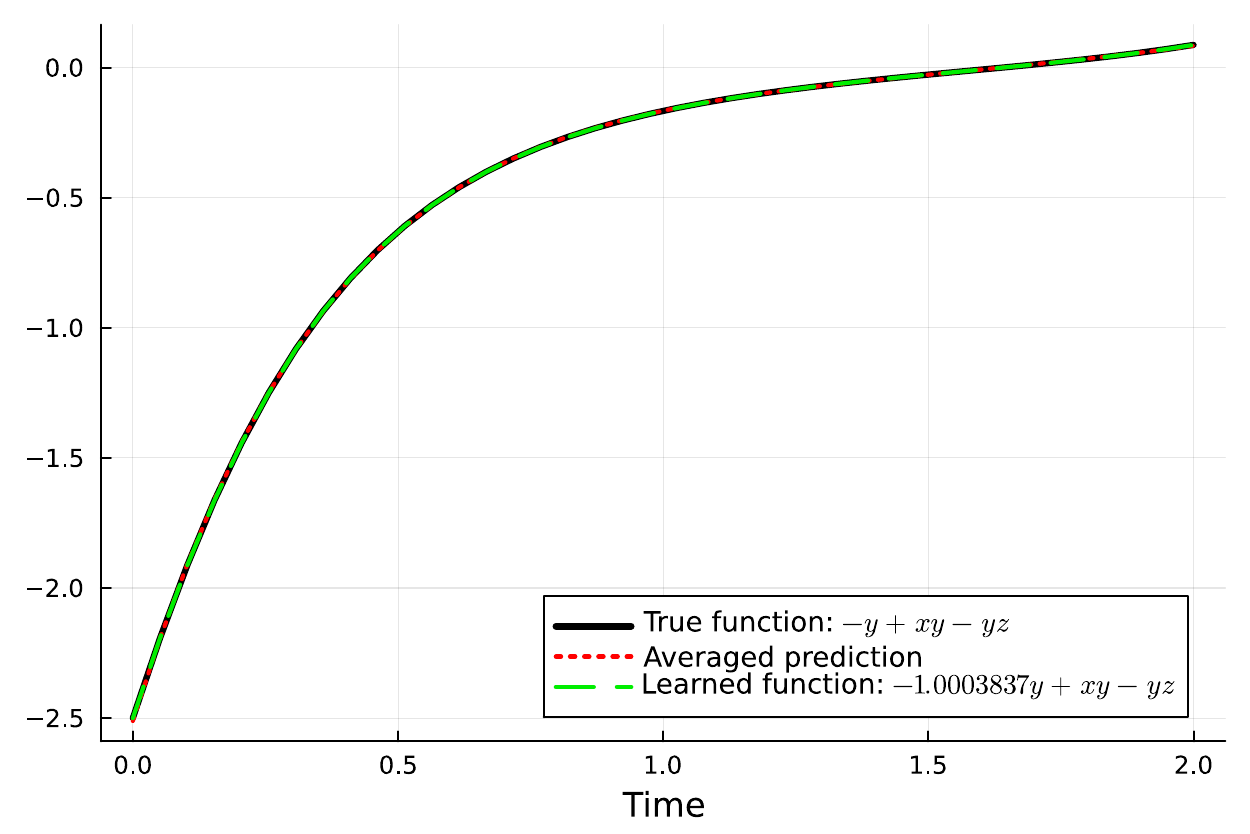}
    \caption{0\% noise}
    \label{fig:LV3AvgDynamics0}
\end{subfigure}
\begin{subfigure}{0.49\linewidth}
\centering
    \includegraphics[width=\linewidth]{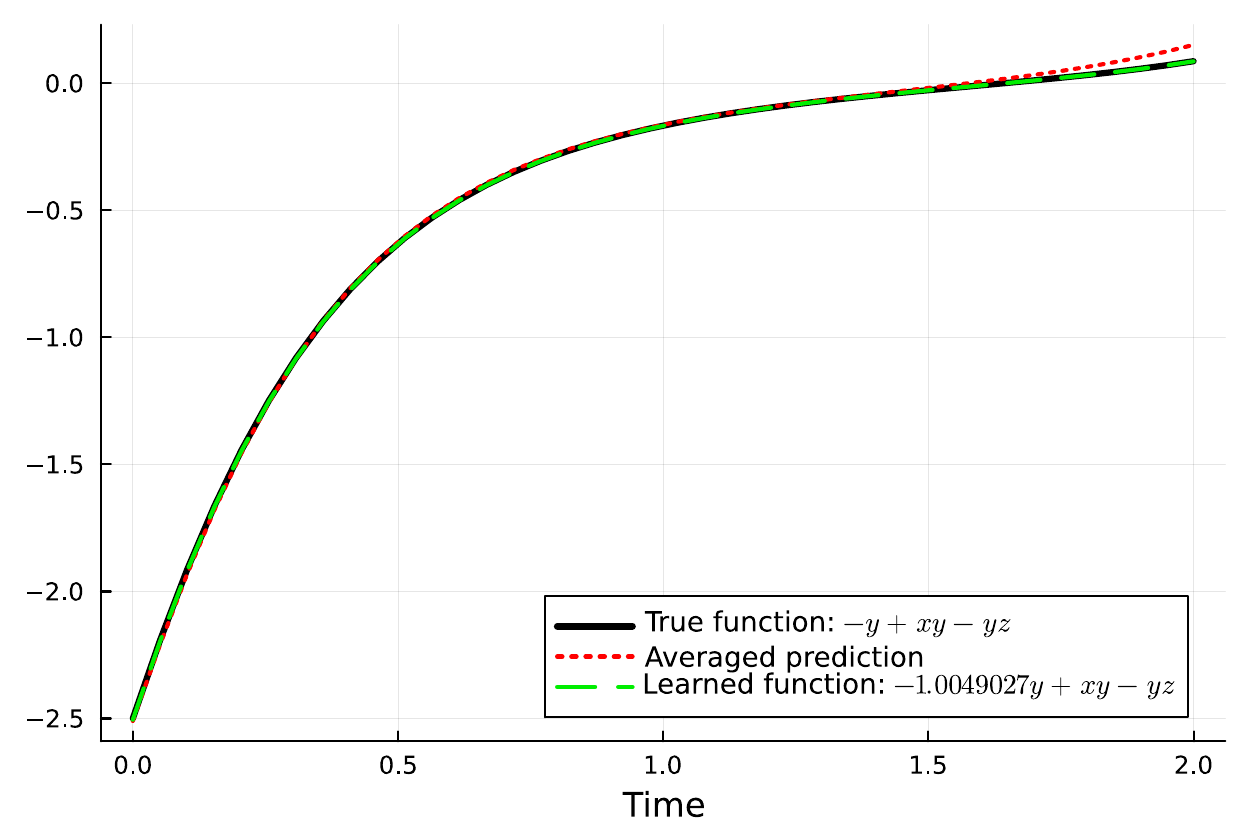}
    \caption{2\% noise}
    \label{fig:LV3AvgDynamics2}
\end{subfigure}
\begin{subfigure}{0.49\linewidth}
\centering
    \includegraphics[width=\linewidth]{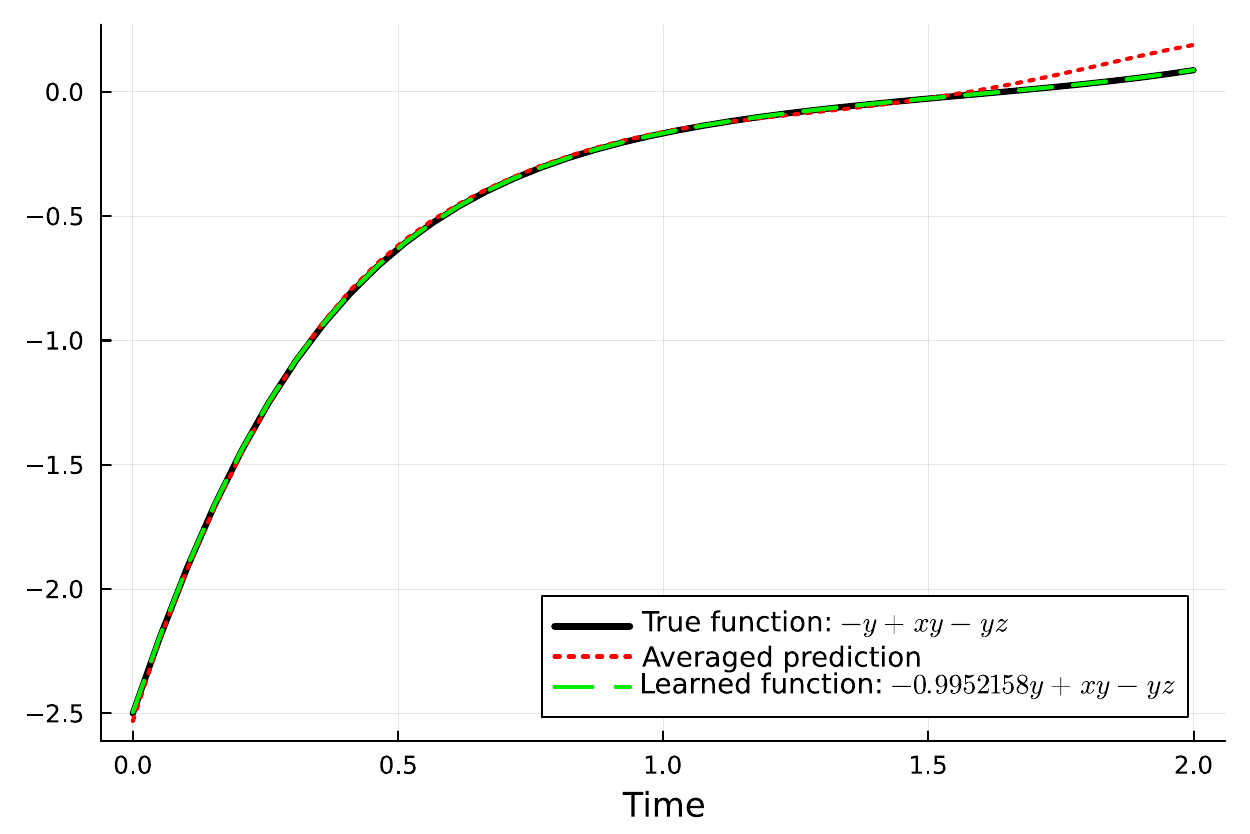}
    \caption{5\% noise}
    \label{fig:LV3AvgDynamics5}
\end{subfigure}
    \caption{The averaged predicted dynamics of the $\frac{dy}{dt}$ equation generated by each of the networks in the ensemble (red) and the corresponding learned function (green), at each level of noise added to the ground truth data.}
    \label{fig:LV3AvgDynamics}
\end{figure}

Figure \ref{fig:LV3Dynamics0} shows that each network generates a very similar and accurate prediction. Therefore, the method of training an ensemble of models and applying SR to the averaged prediction is likely not necessary in the absence of measurement noise, but is still used for consistency. Figures \ref{fig:LV3Dynamics2} and \ref{fig:LV3Dynamics5} show increased variability in the predictions, as expected. Figure \ref{fig:LV3AvgDynamics} shows the averaged predictions of the networks (red dotted curves) and the learned functions through SR (green dashed curves), at each level of noise.

As expected, the averaged prediction in the absence of noise (Figure \ref{fig:LV3AvgDynamics0}) follows the true dynamics very closely. For 2\% and 5\% noise (Figures \ref{fig:LV3AvgDynamics2} and \ref{fig:LV3AvgDynamics5}, respectively), there is a similar deviation of the averaged prediction from the true dynamics towards the end of the training range. Despite this, SR is able to recover the equation with the correct symbolic form due to the remainder of the dynamics being accurately approximated by the averaged prediction. Table \ref{tab:LV3LearnedFunctions} shows the equations discovered by SR at each level of noise.

\begin{table}[H]
\centering
\footnotesize
\begin{tabular}{| c | c | c |}\hline
\textbf{Noise} & \textbf{Learned Equation} & \textbf{True Equation}\\ \hline
0\% & $\dot{y}=-1.0003837y + xy - yz$ & \multirow{3}{2.2cm}{$\dot{y}=-y+xy-yz$}\\
2\% & $\dot{y}=-1.0049027y + xy - yz$ &\\
5\% & $\dot{y}=-0.9952158y + xy - yz$ &\\
\hline
\end{tabular}
\caption{\label{tab:LV3LearnedFunctions} Learned equations for $\frac{dy}{dt}$ at each level of noise added to the ground truth data. The target data for SR was the averaged prediction of the ensemble (red curves in Figure \ref{fig:LV3AvgDynamics}).}
\end{table}

Table \ref{tab:LV3LearnedFunctions} shows that at each level of noise, the correct equation structure is discovered, with the only difference being the coefficient of the $y$ term. The learned equation in the absence of noise is then substituted for the neural network in the hybrid neural ODE, resulting in a partially-learned system, where equation \ref{1.2} is set to be $\frac{dy}{dt}=-1.0003837y + xy - yz$.

\begin{figure}[H]
    \centering
    \includegraphics[width=0.8\linewidth]{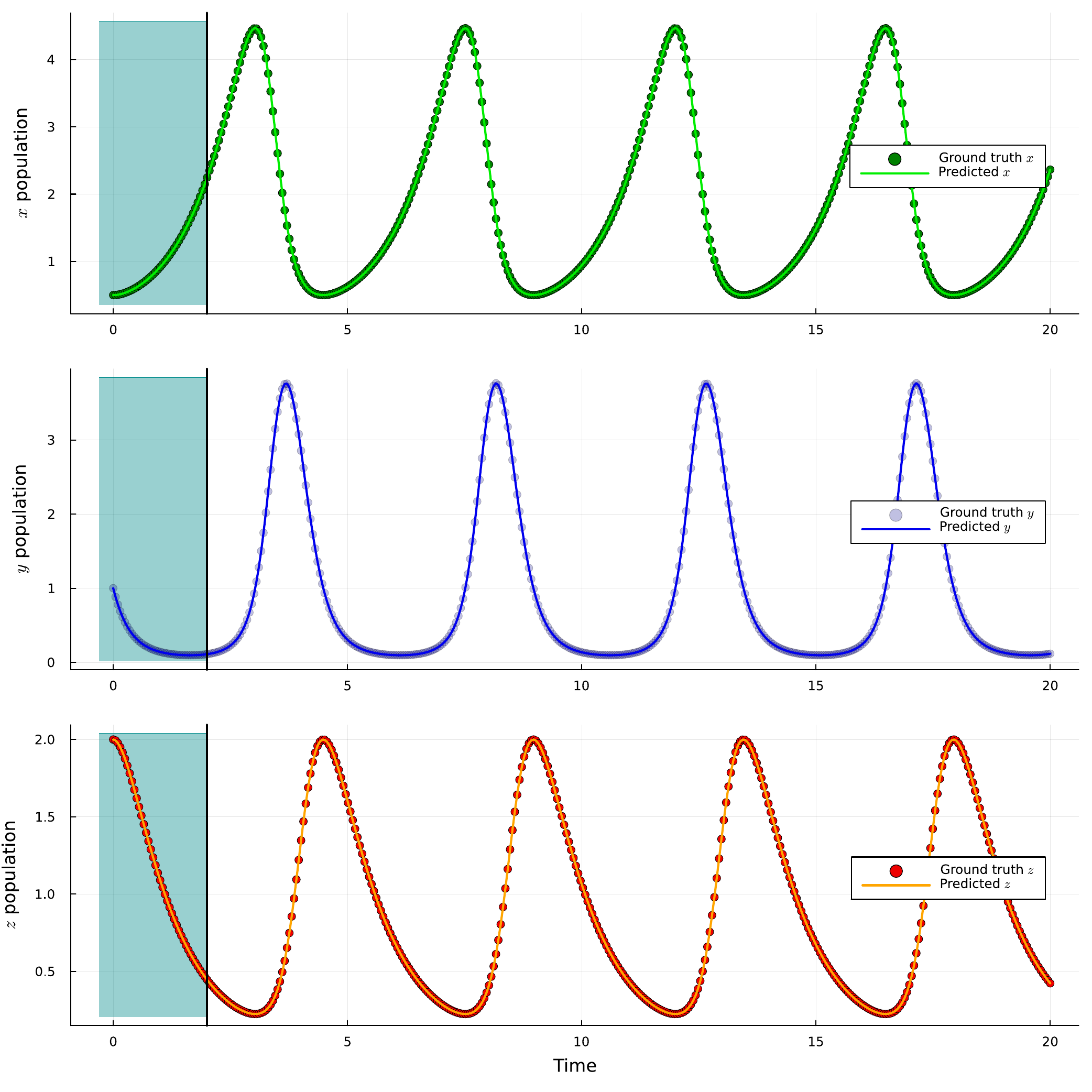}
    \caption{Extrapolation of the partially-learned model, where $\frac{dy}{dt}=-1.0003837y + xy - yz$. The predictions are shown by the solid curves and the ground truth is shown by the scatter points. The ground truth for the state variable $y$ is faint as this data is assumed to be unavailable. The blue region to the left of each plot represents the training range.}
    \label{fig:LV3LearnedEx0}
\end{figure}

To examine this partially-learned model's ability to generalise, another extrapolation up to 20 units of time is made. The resulting predictions are shown in Figure \ref{fig:LV3LearnedEx0}, which accurately track the dynamics of the true system.

\subsection{5D Lorenz System} \label{sec:5DLorenzResults}

Figures show the predictions of one of the ten hybrid neural ODEs from the ensemble before and after the training process, respectively. In this case, the $x$ and $y$ state variables are faint as these data are not provided to the model in training.

\begin{figure}[H]
    \centering
\begin{subfigure}{0.49\linewidth}
\centering
    \includegraphics[width=\linewidth]{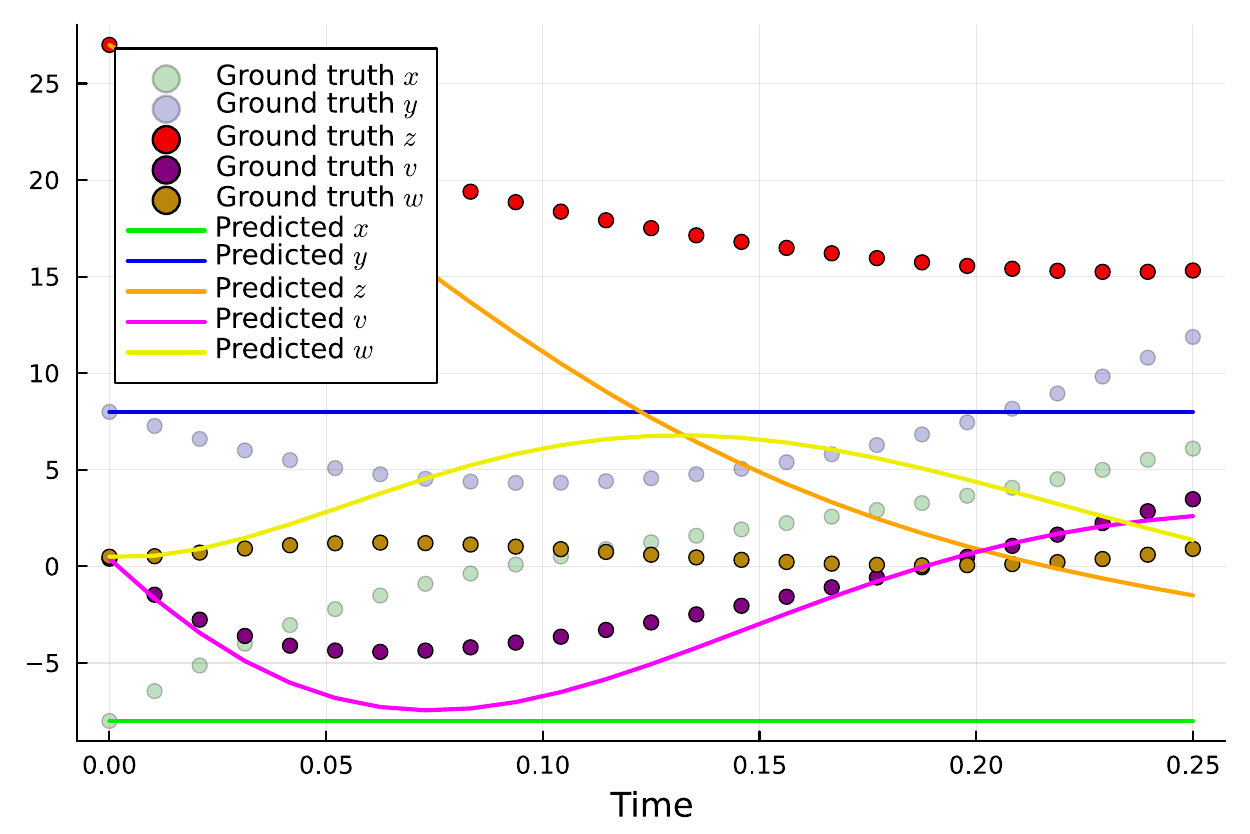}
    \caption{Untrained hybrid neural ODE prediction}
    \label{fig:Lorenz5DUntrained}
\end{subfigure}
\begin{subfigure}{0.49\linewidth}
\centering
    \includegraphics[width=\linewidth]{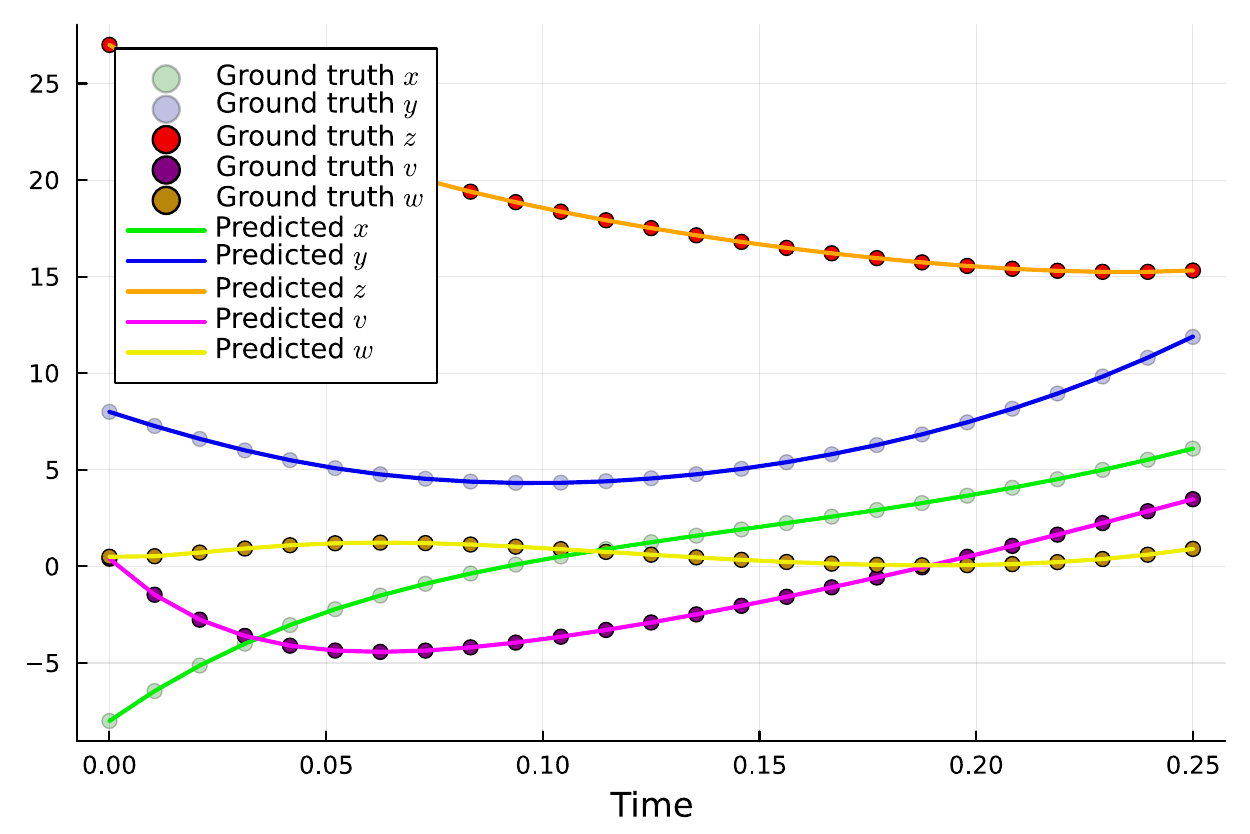}
    \caption{Trained hybrid neural ODE prediction}
    \label{fig:Lorenz5DTrained}
\end{subfigure}
    \caption{Prediction of the hybrid neural ODE (shown by the solid curves) against the ground truth data (shown by the scatter points), both before (Figure \ref{fig:Lorenz5DUntrained}) and after (Figure \ref{fig:Lorenz5DTrained}) the training process. The ground truth data for the state variables $x$ and $y$ are faint as this data is not available during training.}
    \label{fig:Lorenz5DTraining}
\end{figure}

\begin{figure}[H]
    \centering
    \begin{subfigure}{0.49\linewidth}
    \centering
    \includegraphics[width=\linewidth]{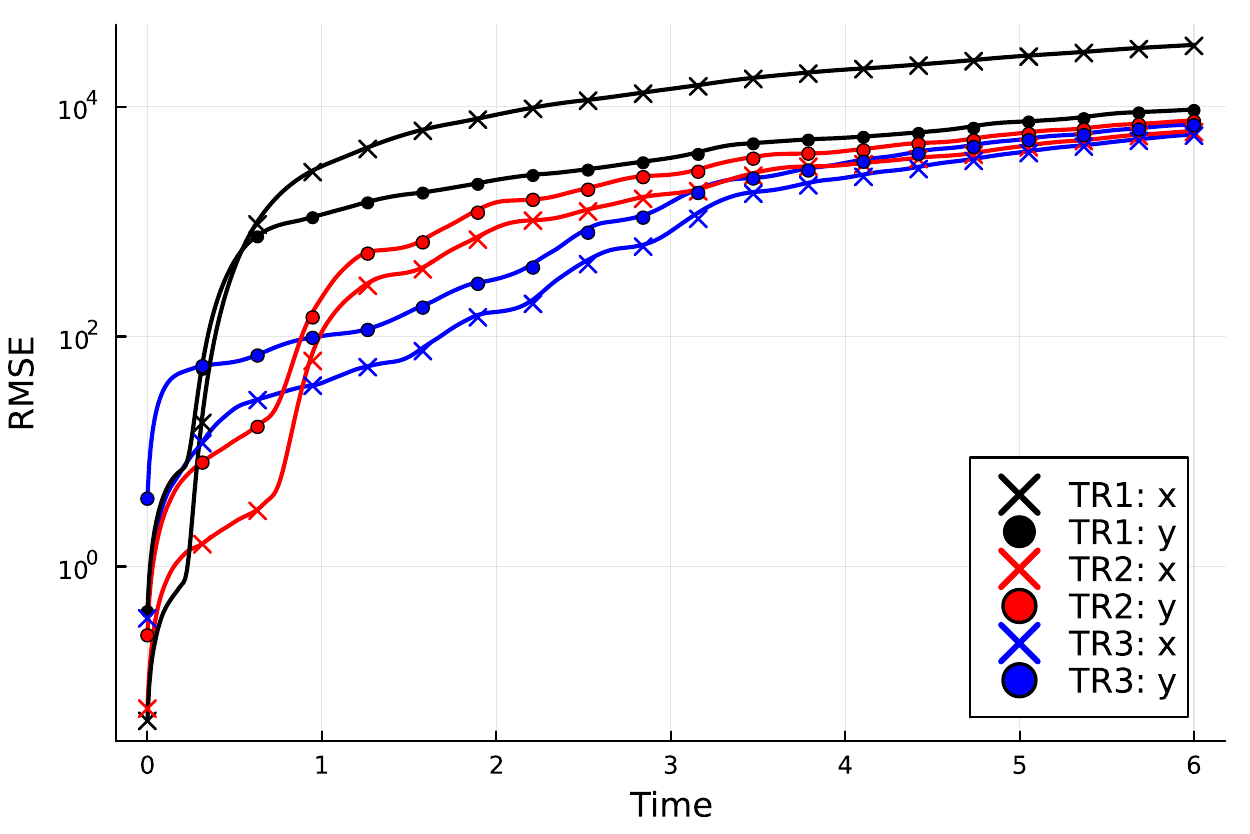}
    \label{fig:Lorenz5DErrors0noise}
    \caption{0\% noise}
    \end{subfigure}
    \begin{subfigure}{0.49\linewidth}
    \centering
    \includegraphics[width=\linewidth]{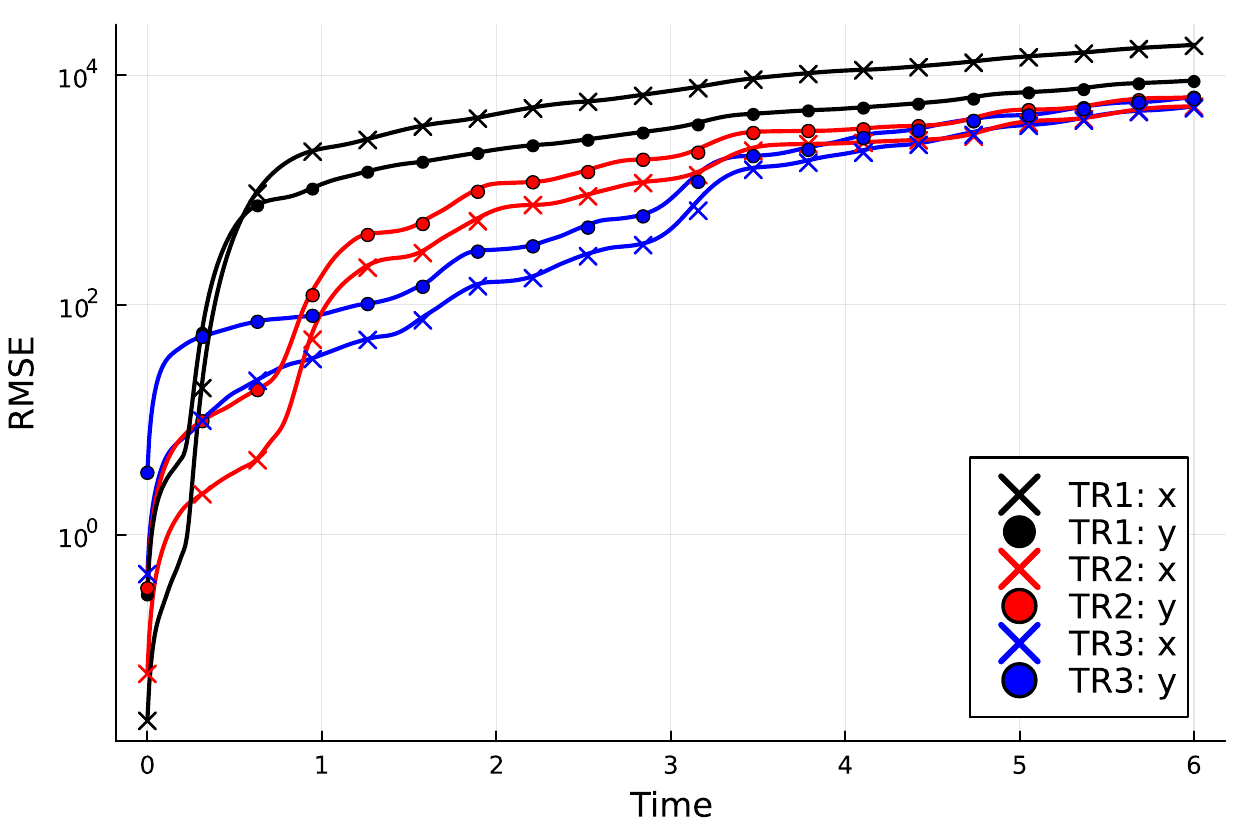}
    \label{fig:Lorenz5DErrors1noise}
    \caption{0.3\% noise}
    \end{subfigure}
    \caption{Sliding window errors of hybrid neural ODE extrapolations, using three different training ranges. The training ranges of 0.4, 0.8 and 1.2 units of time are abbreviated to TR1, TR2 and TR3, respectively. Only the unobserved states are shown, with $x$ and $y$ being represented by the crosses and circles, respectively. This comparison is done at each level of noise.}
    \label{fig:Lorenz5DErrors}
\end{figure}

Figure \ref{fig:Lorenz5DTraining} shows that the hybrid neural ODE is able to accurately capture the dynamics of the system within the training range, for both the observed and unobserved states. To investigate the model's ability to generalise beyond the training data, an extrapolation up to 6 units of time is made. However, the short training range means that even with the physical knowledge available through the $\frac{dz}{dt}, \frac{dv}{dt}$ and $\frac{dw}{dt}$ equations, there is insufficient information available for the network to produce accurate extrapolations. This, together with chaotic nature of the Lorenz attractor, results in an unstable prediction and is hence not shown. It is for this reason that the training range of 0.25 units of time is omitted from the comparative analysis of the effects of varying the length of the training range on the accuracy of the extrapolations (0.4, 0.8 and 1.2 units of time are used for this comparison). Figure \ref{fig:Lorenz5DErrors} shows the results of this analysis. Increasing the training range improves the accuracy of the extrapolation, however the chaotic nature of the system ensures a divergence from the ground truth relatively quickly, even in the case of the longest training range. As a result, the errors for each training range converge towards the end of the simulation.

\begin{figure}[H]
    \centering
\begin{subfigure}{\linewidth}
\centering
    \includegraphics[width=0.49\linewidth]{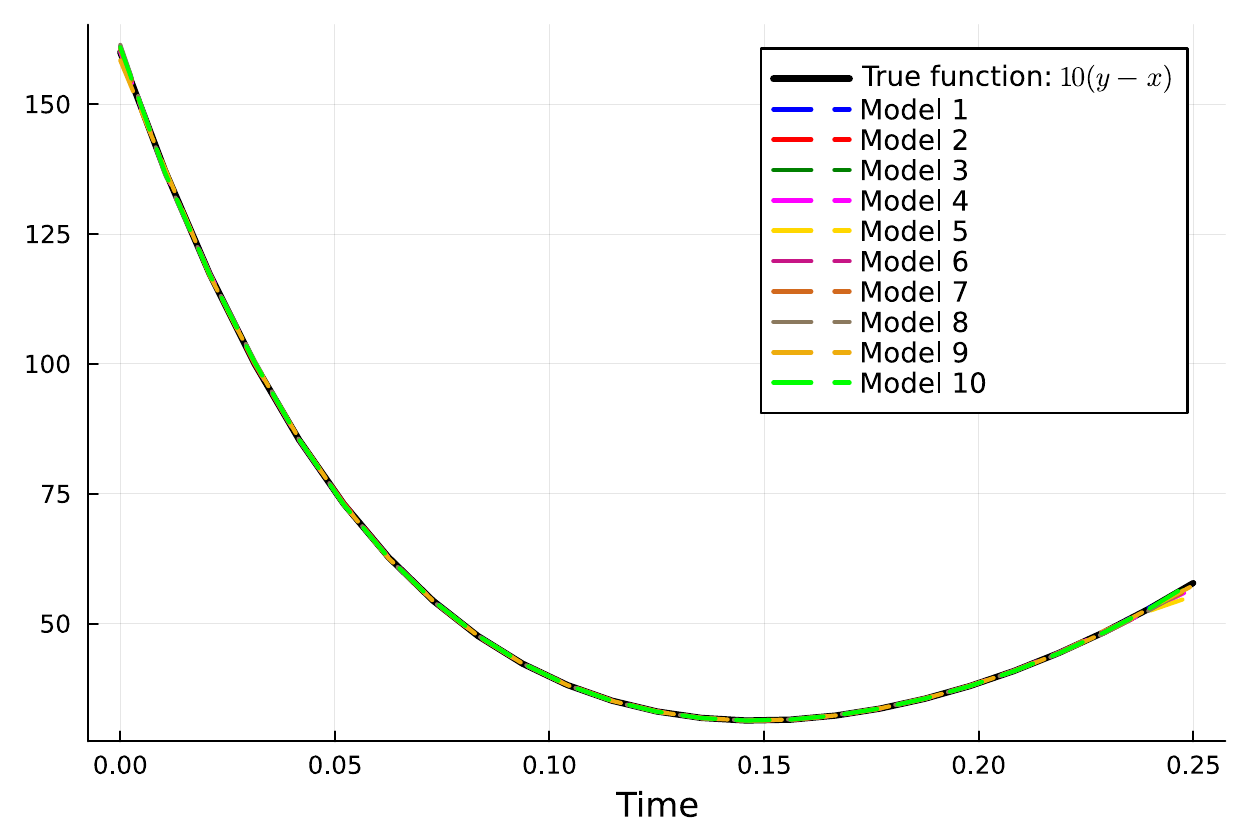}
    \includegraphics[width=0.49\linewidth]{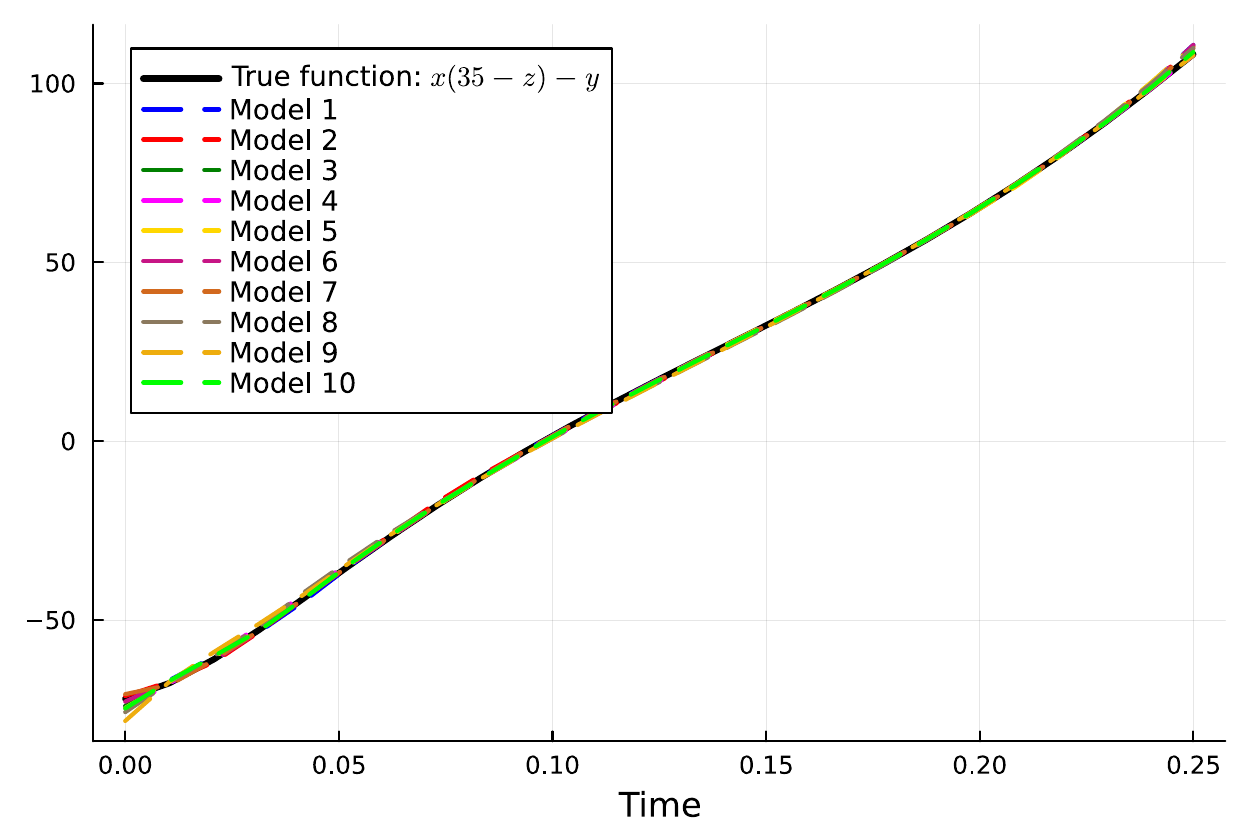}
    \caption{0\% noise}
    \label{fig:Lorenz5DDynamics0}
\end{subfigure}
\begin{subfigure}{\linewidth}
\vspace{0.5cm}
\centering
    \includegraphics[width=0.49\linewidth]{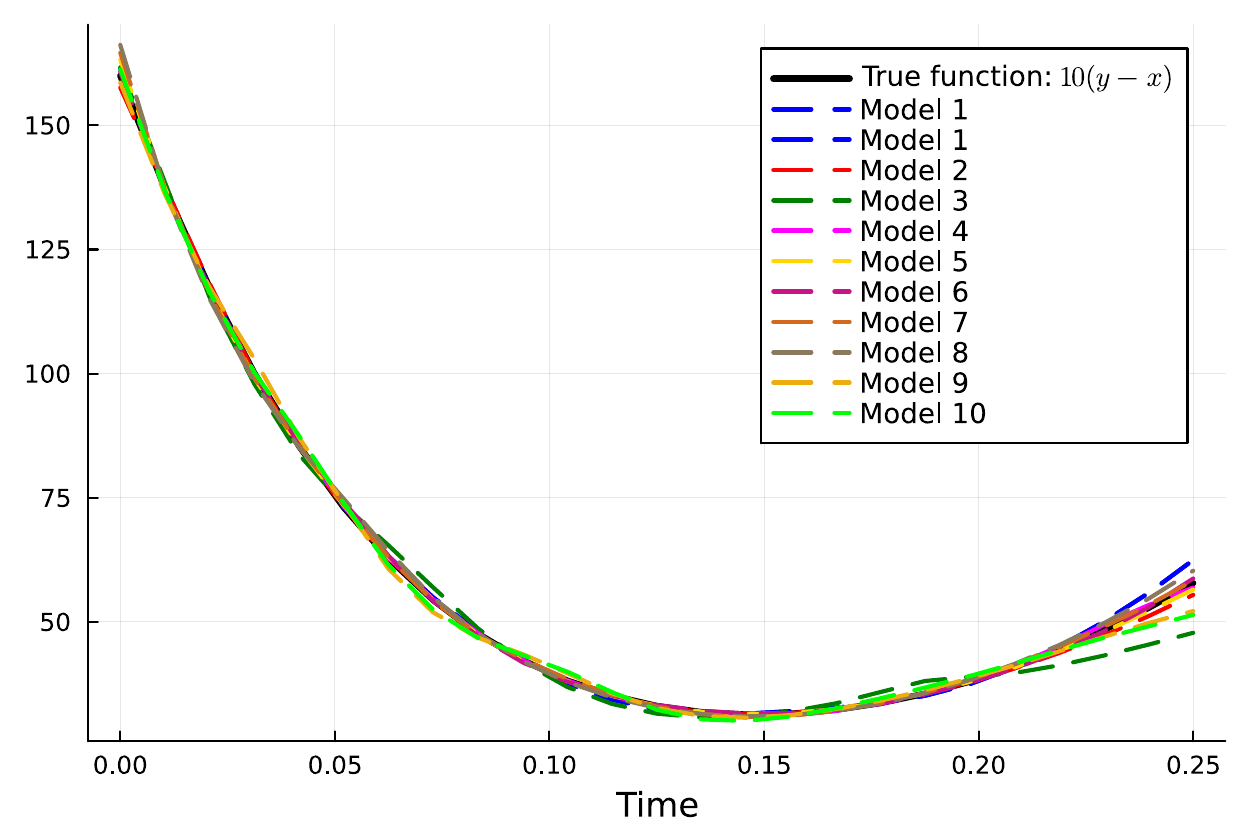}
    \includegraphics[width=0.49\linewidth]{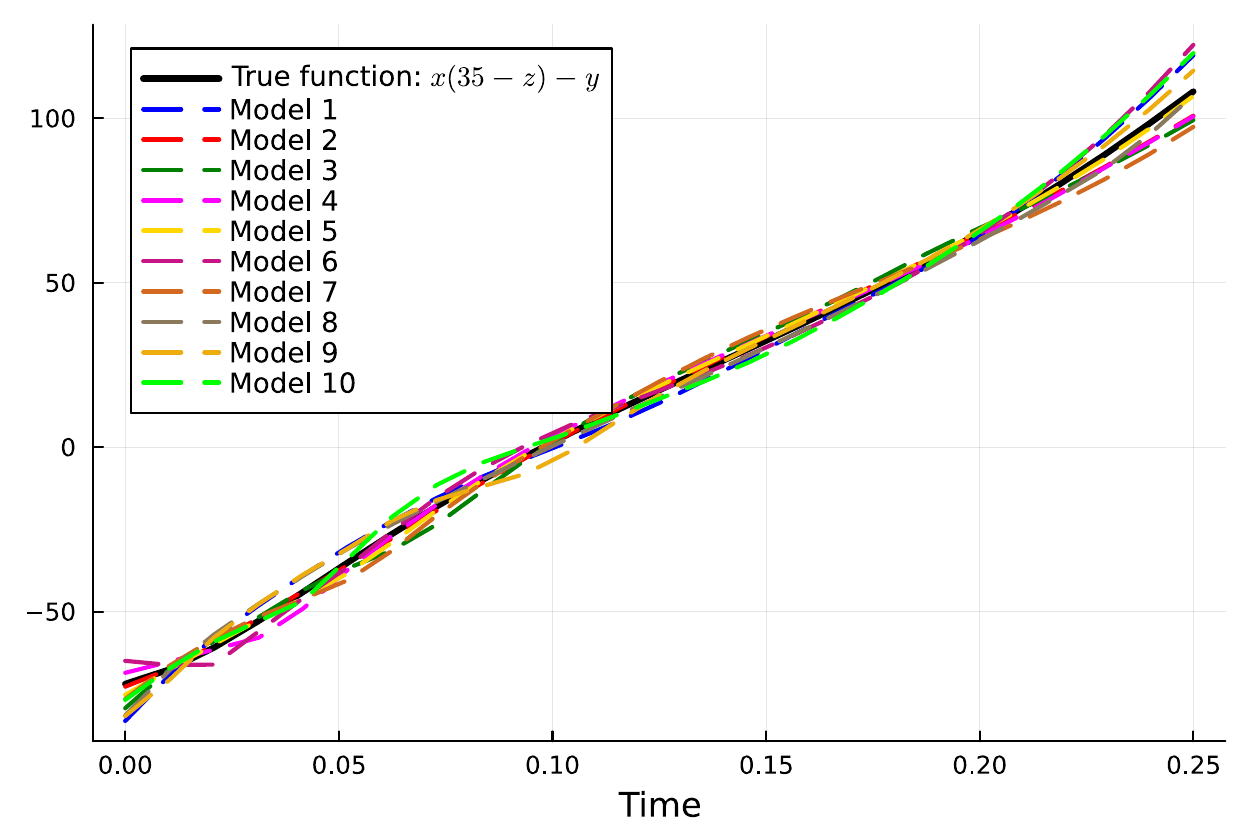}
    \caption{0.3\% noise}
    \label{fig:Lorenz5DDynamics3}
\end{subfigure}
    \caption{The predicted dynamics of the $\frac{dx}{dt}$ and $\frac{dy}{dt}$ equations generated by each network in the ensemble, at each level of noise added to the ground truth data.}
    \label{fig:Lorenz5DDynamics}
\end{figure}

The issue of a relatively quick divergence from the ground truth can be ameliorated if the correct equations are discovered by SR. Returning to the case of a training range of 0.25 units of time, the ten hybrid neural ODEs in the ensemble are all simulated to generate the temporal evolution of the states $[x,y,z,v,w]$, which are then passed as input to each of the ten neural networks. The outputs of each network then correspond to the learned dynamics of the $\frac{dx}{dt}$ and $\frac{dy}{dt}$ equations. This is shown in Figure \ref{fig:Lorenz5DDynamics}.

\begin{figure}[H]
    \centering
\begin{subfigure}{\linewidth}
\centering
    \includegraphics[width=0.49\linewidth]{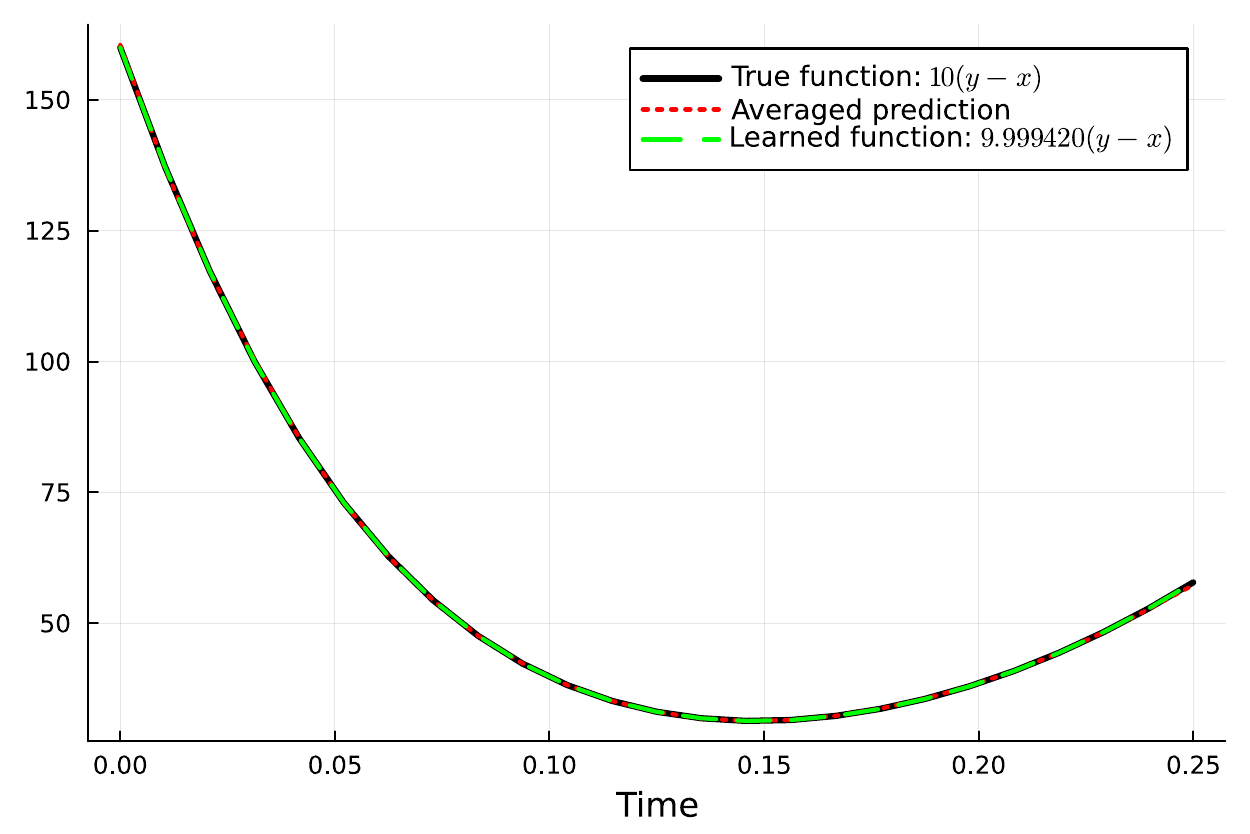}
    \includegraphics[width=0.49\linewidth]{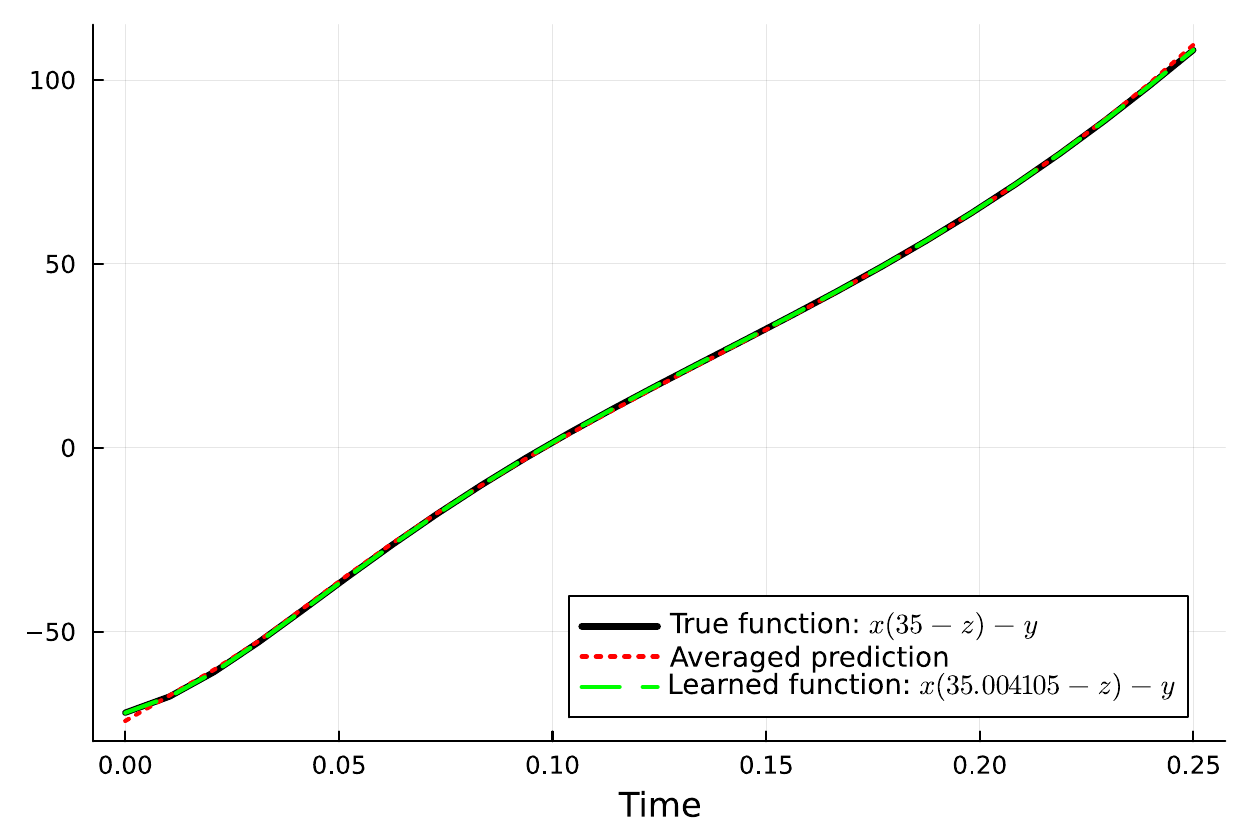}
    \caption{0\% noise}
    \label{fig:Lorenz5DAvgDynamics0}
\end{subfigure}
\begin{subfigure}{\linewidth}
\vspace{0.5cm}
\centering
    \includegraphics[width=0.49\linewidth]{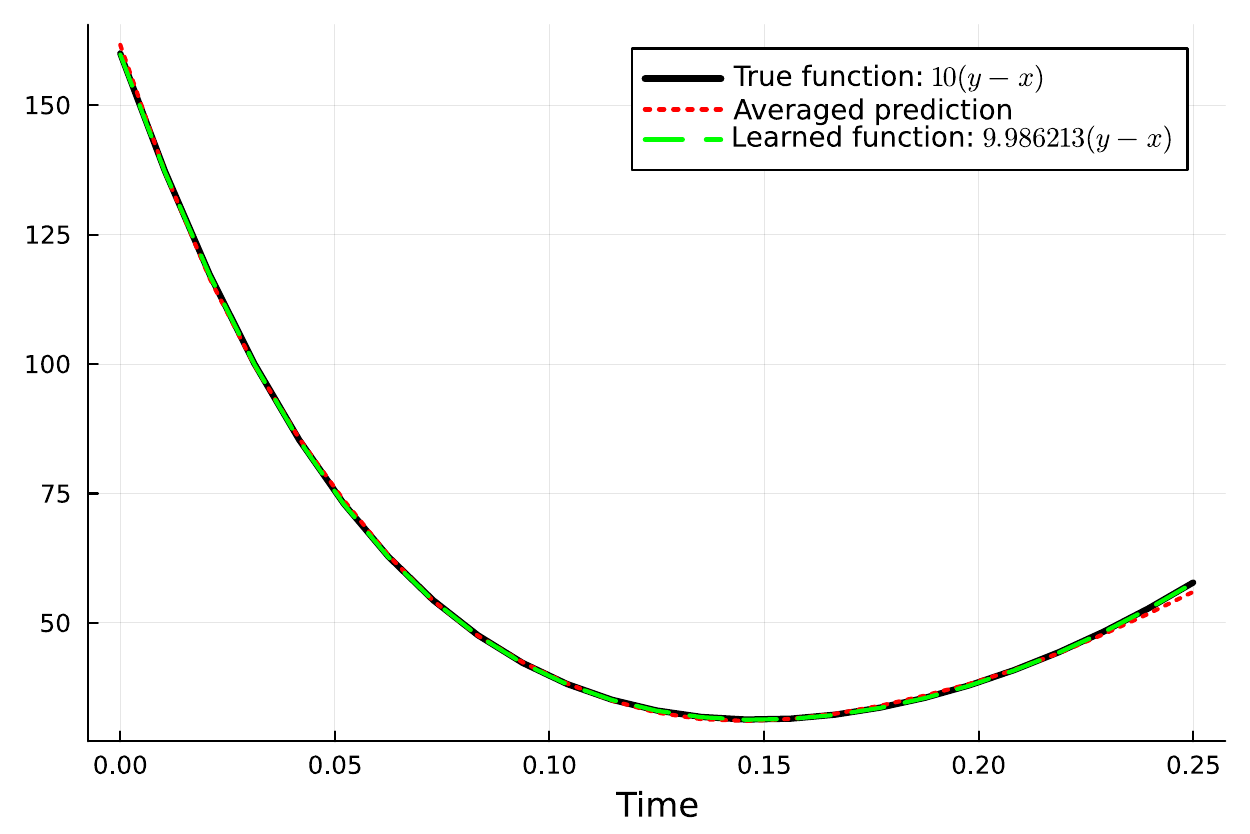}
    \includegraphics[width=0.49\linewidth]{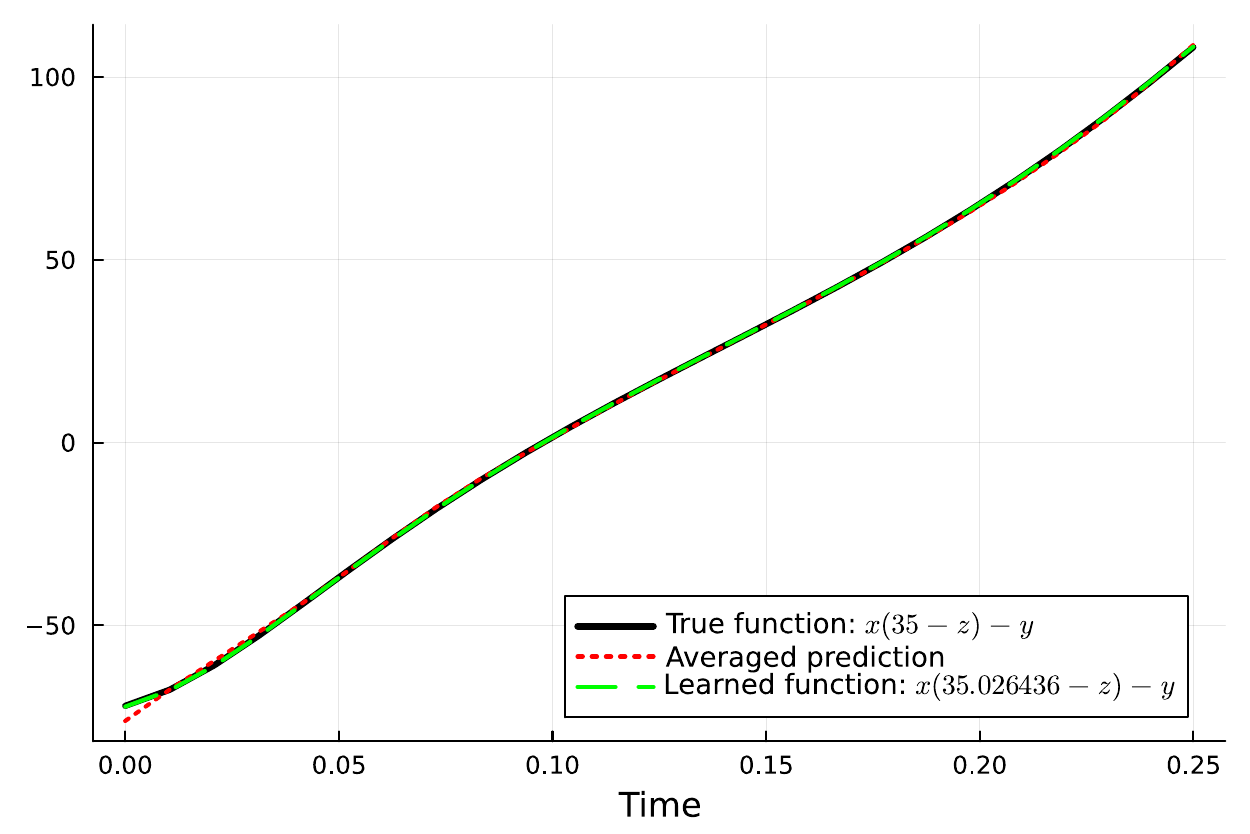}
    \caption{0.3\% noise}
    \label{fig:Lorenz5DAvgDynamics3}
\end{subfigure}
    \caption{The averaged predicted dynamics of the $\frac{dx}{dt}$ and $\frac{dy}{dt}$ equations generated by each of the networks in the ensemble (red) and the corresponding learned function (green), at each level of noise added to the ground truth data.}
    \label{fig:Lorenz5DAvgDynamics}
\end{figure}

Figure \ref{fig:Lorenz5DDynamics0} shows little variability among the predictions of the networks in the ensemble for both the  $\frac{dx}{dt}$ and  $\frac{dy}{dt}$ equations. This variability increases significantly with a small amount of measurement noise added to the training data, as in Figure \ref{fig:Lorenz5DDynamics3}. SR is then applied to the average of these predictions. This is displayed in Figure \ref{fig:Lorenz5DAvgDynamics}, which shows the averaged prediction (red dotted curve) along with the corresponding learned function through SR (green dashed curve).

Figure \ref{fig:Lorenz5DAvgDynamics} highlights the benefit of training an ensemble of models and averaging the network predictions, as the errors cancel out, producing a curve which more accurately approximates the dynamics of the true equations \ref{1.4} and \ref{1.5}. Despite small deviations of the averaged prediction from the true dynamics, SR is able to recover true equation structure. The learned equations are given in Table \ref{tab:Lorenz5DLearnedFunctions}.

\begin{table}[H]
\centering
\footnotesize
\begin{tabular}{| c | c | c |}\hline
\textbf{Noise} & \textbf{Learned Equations} & \textbf{True Equations}\\ \hline
\multirow{2}{0.7cm}{0\%} & $\dot{x}=9.999420(y-x)$ & \multirow{4}{2.2cm}{$\dot{x}=10(y-x)$\\ $\dot{y}=x(35-z)-y$} \\
& $\dot{y}=x(35.004105-z)-y$ &\\
\cline{1-2}
\multirow{2}{0.7cm}{0.3\%} & $\dot{x}=9.986213(y-x)$ &\\
& $\dot{y}=x(35.026436-z)-y$ &\\
\hline
\end{tabular}
\caption{\label{tab:Lorenz5DLearnedFunctions} Learned equations for $\frac{dx}{dt}$ and $\frac{dy}{dt}$ at each level of noise added to the ground truth data. The target data for SR was the averaged prediction of the ensemble (red curves in Figure \ref{fig:Lorenz5DAvgDynamics}).}
\end{table}

The learned equations in Table \ref{tab:Lorenz5DLearnedFunctions} are of the correct form, with the only differences from the true equations being   the coefficient of the $y$ term for the $\frac{dx}{dt}$ equation and the coefficient of the $x$ term in the $\frac{dy}{dt}$ equation. The learned equations in the absence of noise are then substituted back into the model, resulting in a partially-learned system, where equations \ref{1.4} and \ref{1.5} are set to be $\frac{dx}{dt}=9.999420(y - x)$ and $\frac{dy}{dt}=x(35.004105 - z) - y$, respectively.

\begin{figure}[H]
    \centering
    \includegraphics[width=0.8\linewidth]{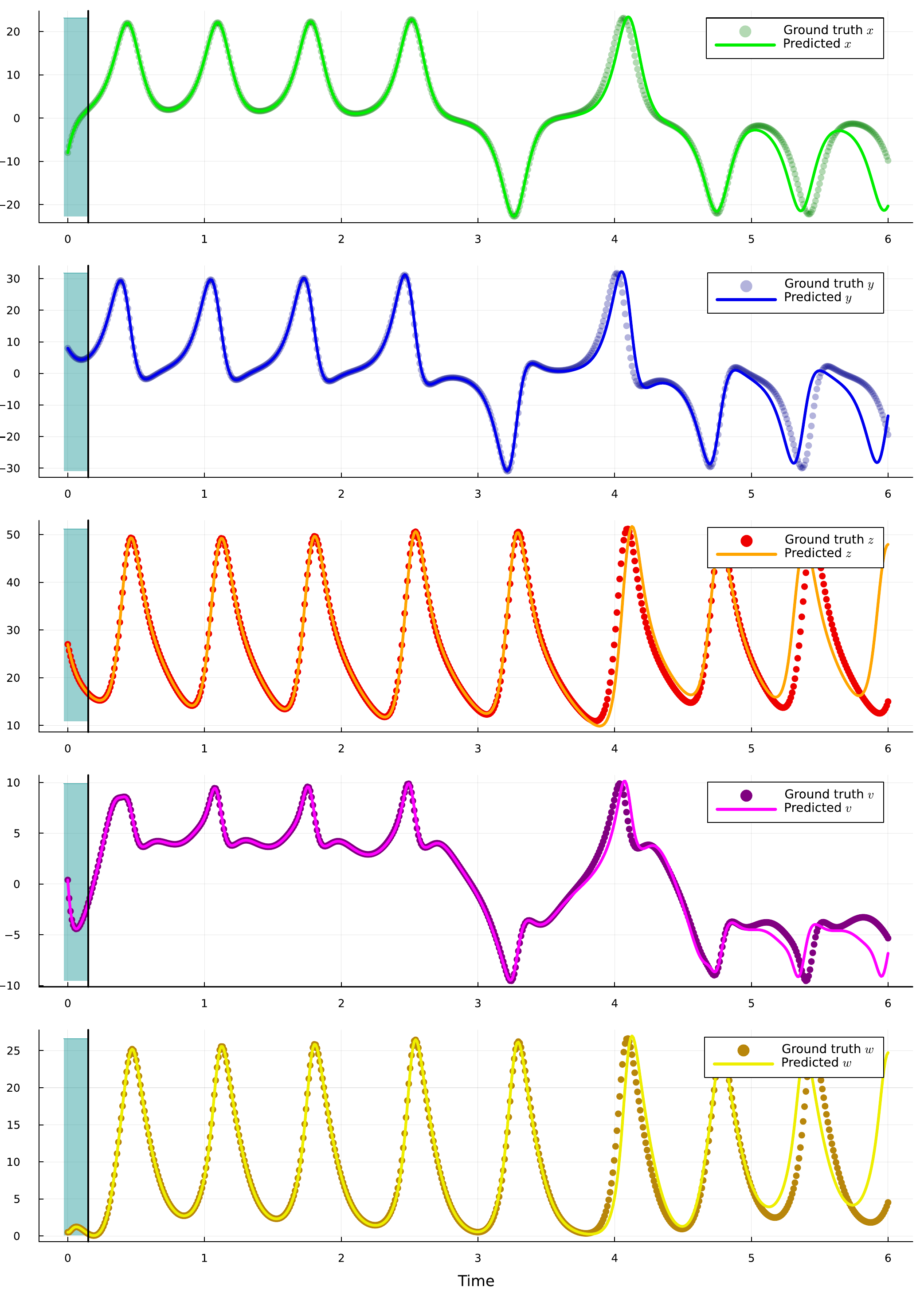}
    \caption{Extrapolation of the partially-learned model with the equations discovered in the absence of noise. Here, $\dot{x}=9.999420(y-x)$ and $\dot{y}=x(35.004105-z)-y$. The predictions are shown by the solid curves and the ground truth is shown by the scatter points. The ground truth for the $x$ and $y$ state variables are faint as this data are assumed to be unavailable. The blue region to the left of each plot represents the training range.}
    \label{fig:Lorenz5DLearnedEx0}
\end{figure}

To assess this partially-learned model's ability to generalise beyond the training data, an extrapolation up to 6 units of time is made, generating the temporal evolution shown in Figure \ref{fig:Lorenz5DLearnedEx0}.

The predictions of the partially-learned model are accurate up to approximately 5 units of time. The deviation of the predictions from the ground truth beyond 5 units of time is expected given the chaotic behaviour of the Lorenz system in that small perturbations to the parameters of the system have a significant effect.

\section{Discussion and Conclusion}

In this work, we demonstrate that it is possible to recover the underlying governing equations of unobserved states in dynamical systems prescribed by ODEs. The ensemble technique of training multiple hybrid neural ODEs and taking the averaged output of the networks as the target data for SR proved to be an effective means of ensuring that the true governing equations were recovered in the presence of measurement noise.

If data are available on the function to be learned, then SR can be applied directly to the data. However, the cases where data are unavailable motivates the use of the neural network, where the missing dynamics can first be learned by training on the available data, before applying SR.

In the PySR implementation, a list of the most suitable candidate functions is generated. Generally, the most complex function (often with many nested terms) has the lowest error. However, this can be an over-fit and can lack interpretability. Therefore, PySR assigns each function a `score', based on a trade-off between accuracy and complexity \cite{cranmer2023interpretable}. In all experiments conducted in this work, the learned equations presented were the ones with the highest score from the list of generated functions.

For both the Lotka-Volterra and Lorenz systems addressed in this work, the training range was selected to be relatively short, capturing only a small region of the dynamics. While this can often highlight the advantage of incorporating physical knowledge within the model, in that less data is needed to train the embedded neural network, extrapolations of the trained model can be poor if the range of training data is too short. This was the case for both examples in this work. Despite this, a short training range was used to allow the neural network to approximate the missing dynamics more closely, in order to increase the likelihood of learning the true underlying equations via SR.

For the hybrid neural ODEs that were trained on noisy data, the extrapolations of the corresponding partially-learned models are shown in Appendix \ref{Appendix}. For the Lotka-Volterra system, the partially-learned models constructed from 2\% and 5\% measurement noise generate predictions (Figures \ref{fig:LV3LearnedEx2} and \ref{fig:LV3LearnedEx5}, respectively) very similar to the partially-learned model constructed in the absence of noise (Figure \ref{fig:LV3LearnedEx0}). This is because the dynamics of the system are periodic and the learned $\frac{dy}{dt}$ equations at each level of noise (Table \ref{tab:LV3LearnedFunctions}) are all similar. For this system, the method is generally not robust to 10\% measurement noise. For the 5D Lorenz system, the learned $\frac{dx}{dt}$ and $\frac{dy}{dt}$ equations in the presence of 0.3\% noise (Table \ref{tab:Lorenz5DLearnedFunctions}) are still accurate, but the slight additional error causes a noticeable difference in the extrapolation of the partially-learned model (Figure \ref{fig:Lorenz5DLearnedEx1}), where the predictions are now only accurate up to 3.5 units of time. This is again due to the chaotic nature of the Lorenz system, in that small changes to parameters of the system have large effects at a later stage.

Appendix \ref{Appendix3} details a slightly alternative method, whereby instead of applying SR to the averaged prediction of networks in the ensemble, SR is applied to each network prediction and an average of the learned models is taken. This highlights the effects of the added measurement noise, since more noise generally causes increased variability among the predictions of the networks in the ensemble, which in turn results in increased variability among the learned functions via SR. This approach shows that for the learned terms spanning all learned functions from the ensemble, the true coefficient values generally lie within 1 standard deviation of the mean learned coefficient values.

A possible avenue of future work is to use a Bayesian neural network within the hybrid neural ODE. This way, the training process would only be carried out once (instead of training an ensemble of 10 models) and multiple predictions of the hybrid model could be generated by sampling from the posterior distribution of the network parameters. As a result of this, SR can be carried out for each sample, and a distribution of partially-learned models can also be generated.\\

\textbf{CRediT authorship contribution statement}\\
\textbf{Gevik Grigorian:} Conceptualization, Formal analysis, Investigation, Methodology, Software, Validation, Visualization, Writing - original draft, Writing - review \& editing. \textbf{Sandip V. George:} Conceptualization, Formal analysis, Investigation, Methodology, Software, Supervision, Validation, Visualization, Writing - original draft, Writing - review \& editing. \textbf{Simon Arridge:1} Conceptualization, Methodology, Software, Supervision, Validation, Visualization, Writing - original draft, Writing - review \& editing.\\

\textbf{Data availability}\\
No data was used for the research described in this article.\\

\textbf{Declaration of competing interest}\\
The authors declare that they have no known competing financial interests or personal relationships that could have appeared to influence the work reported in this paper.\\

\textbf{Acknowledgements}\\
This work was undertaken as part of the EPSRC-Funded CHIMERA (\textbf{C}ollaborative \textbf{H}ealthcare \textbf{I}nnovation through \textbf{M}athematics,
\textbf{E}nginee\textbf{R}ing and \textbf{A}I) Mathematical Sciences in Healthcare Hub (grant no. EP/T017791/1), which aims to improve outcomes of critically unwell intensive care unit (ICU) patients through combinations of mechanistic modelling, statistical and machine learning.

\appendix

\section{Appendix: Learned Extrapolations}
\label{Appendix}
The extrapolations of the learned models which were trained on noisy data are given here.

\subsection{3D Lotka-Volterra System}

\begin{figure}[H]
    \centering
    \includegraphics[width=0.8\linewidth]{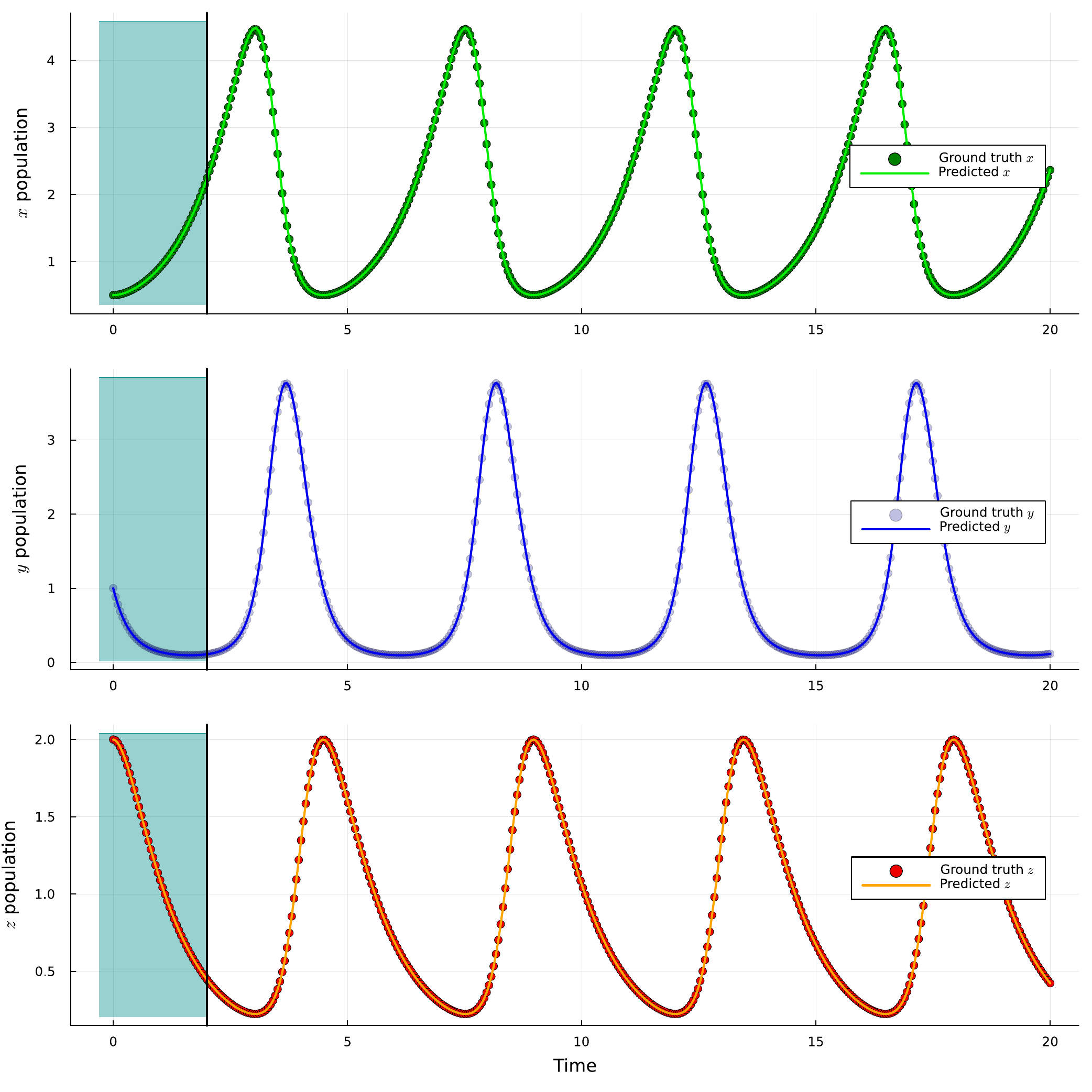}
    \caption{Extrapolation of the partially-learned model with the equations discovered in the presence of 2\% noise. Here, $\dot{y}=-1.0049027y + xy - yz$. The predictions are shown by the solid curves and the ground truth is shown by the scatter points. The ground truth for the state variable $y$ is faint as this data is assumed to be unavailable. The blue region to the left of each plot represents the training range.}
    \label{fig:LV3LearnedEx2}
\end{figure}

Figure \ref{fig:LV3LearnedEx2} shows the extrapolation of the model including the equations learned in the presence of 2\% noise, from Table \ref{tab:LV3LearnedFunctions}.

\begin{figure}[H]
    \centering
    \includegraphics[width=0.8\linewidth]{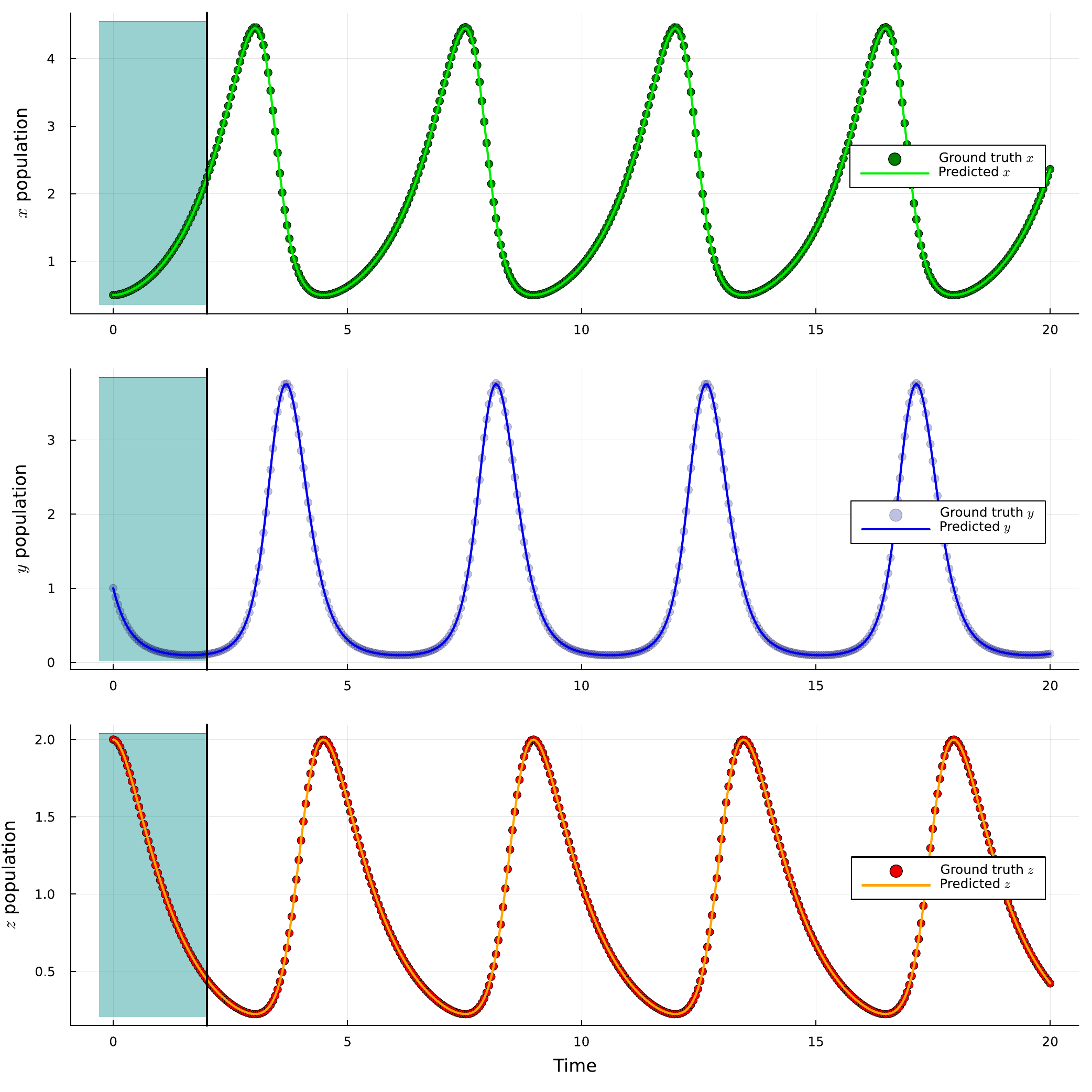}
    \caption{Extrapolation of the partially-learned model with the equations discovered in the presence of 5\% noise. Here, $\dot{y}= -0.9952158y + xy - yz$. The predictions are shown by the solid curves and the ground truth is shown by the scatter points. The ground truth for the state variable $y$ is faint as this data is assumed to be unavailable. The blue region to the left of each plot represents the training range.}
    \label{fig:LV3LearnedEx5}
\end{figure}

Figure \ref{fig:LV3LearnedEx5} shows the extrapolation of the model including the equations learned in the presence of 5\% noise, from Table \ref{tab:LV3LearnedFunctions}.

\subsection{5D Lorenz System}

\begin{figure}[H]
    \centering
    \includegraphics[width=0.8\linewidth]{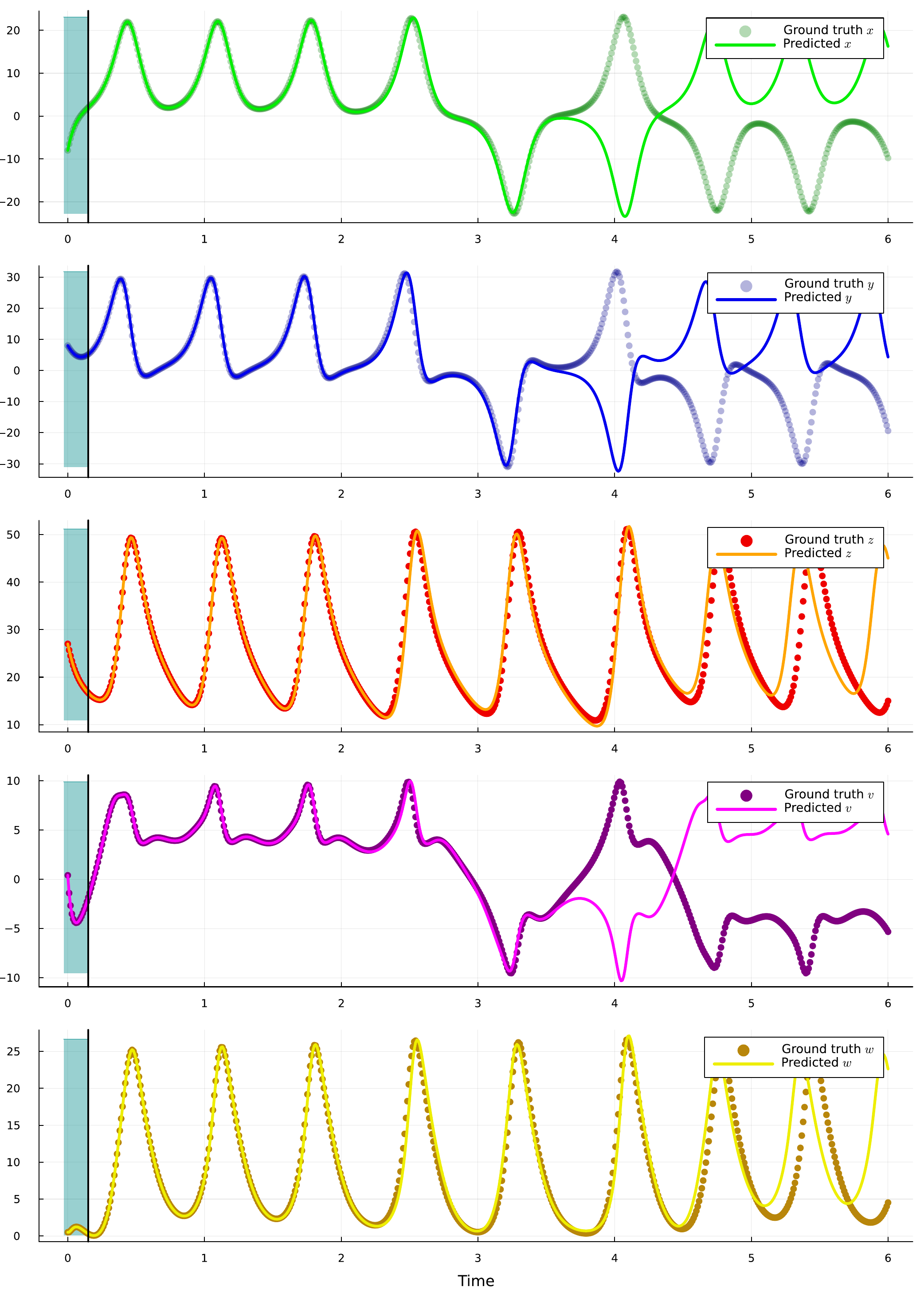}
    \caption{Extrapolation of the partially-learned model with the equations discovered in the presence of 0.3\% noise. Here, $\dot{x}= 9.986213(y - x)$ and $\dot{y}=x(35.026436 - z) - y$. The predictions are shown by the solid curves and the ground truth is shown by the scatter points. The ground truth for the $x$ and $y$ state variables are faint as this data are assumed to be unavailable. The blue region to the left of each plot represents the training range.}
    \label{fig:Lorenz5DLearnedEx1}
\end{figure}

Figure \ref{fig:Lorenz5DLearnedEx1} shows the extrapolation of the model including the equations learned in the presence of 0.3\% noise, from Table \ref{tab:Lorenz5DLearnedFunctions}.

\section{Appendix: PySR Details} 
\label{Appendix2}

The user defined hyper-parameters for the PySR implementation are given in Table \ref{tab:PySRdetails}. The process begins with 1000 sub-populations containing 33 randomly initialised trees each. Via the processes mentioned in section \ref{SymbolicRegression}, new trees are iteratively generated. MSE is selected as the fitness metric to evaluate how well a candidate function fits the target data. Though we select 200 iterations as the stopping criteria for the algorithm, alternative stopping criteria such as an amount of time elapsed or a desired fitness reached, can be used.

\begin{table}[H]
\footnotesize
\begin{center}
\begin{tabular}{| l | r | }\hline
Unary operators & $\{e\}$ \\
\hline
 Binary operators & $\{+, -, \div, \times \}$\\
  \hline
 Functions per population & $33$\\
 \hline
 Populations & 1000\\ \hline
 Iterations & 200\\
 \hline
 Performance metric & MSE\\
 \hline
\end{tabular}
\end{center}
\caption{\label{tab:PySRdetails} Hyper-parameters used in the PySR implementation.}
\end{table}

\section{Appendix: Alternative Method} 
\label{Appendix3}

\begin{figure}[H]
    \centering
    \begin{subfigure}{0.49\linewidth}
    \centering
    \includegraphics[width=\linewidth]{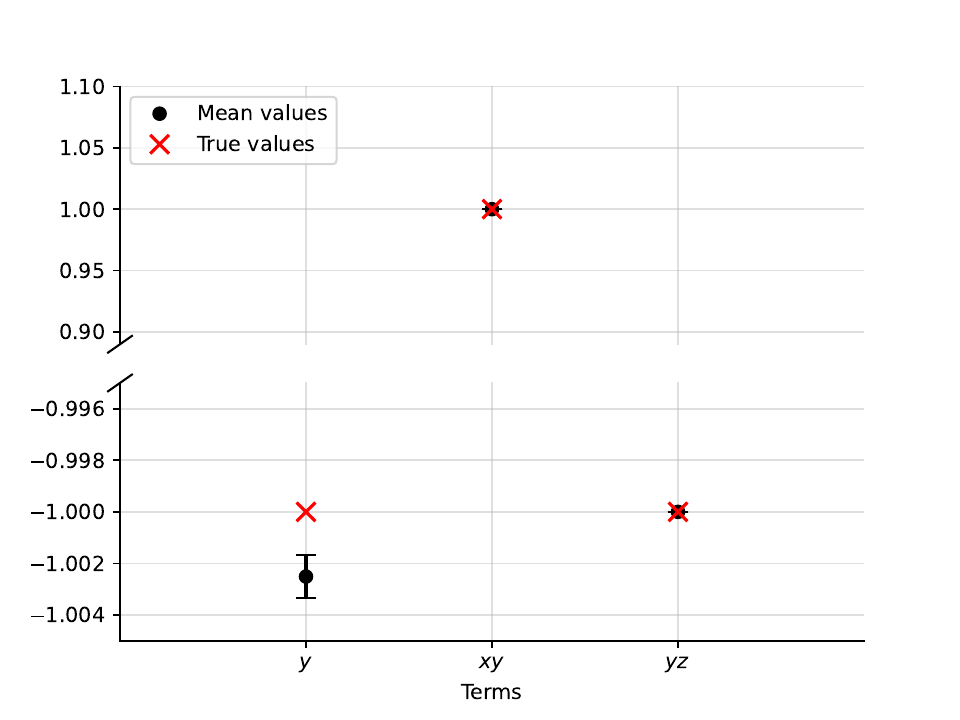}
    \caption{\label{fig:LV3DTerms0Noise} 0\% noise}
    \end{subfigure}
    \begin{subfigure}{0.49\linewidth}
    \centering
    \includegraphics[width=\linewidth]{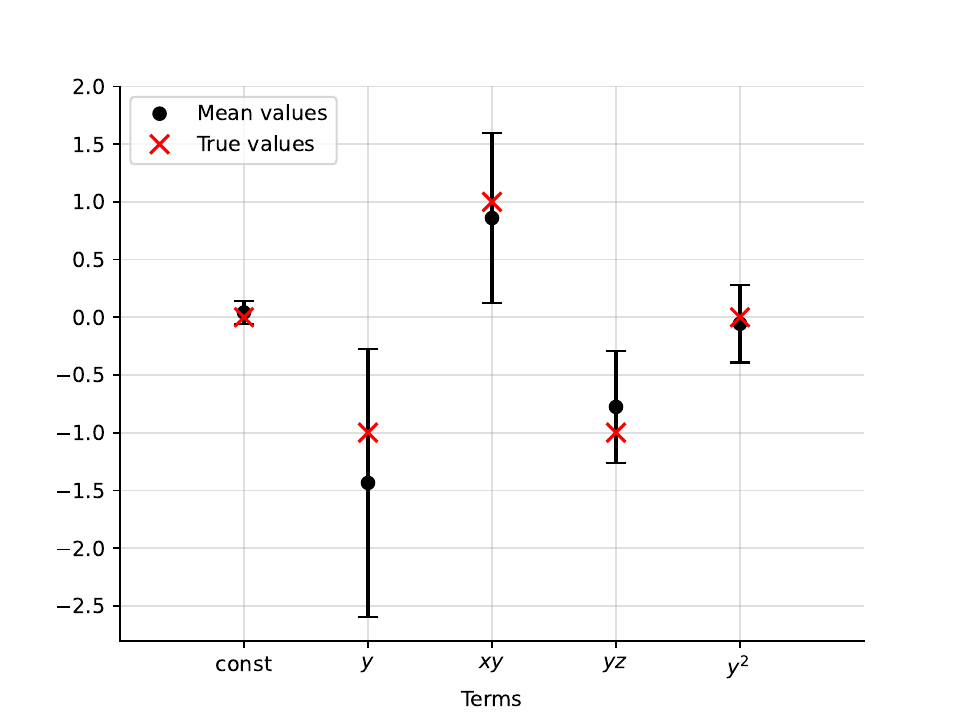}
    \caption{2\% noise \label{fig:LV3DTerms2Noise}}
    \end{subfigure}
    \begin{subfigure}{0.49\linewidth}
    \centering
    \includegraphics[width=\linewidth]{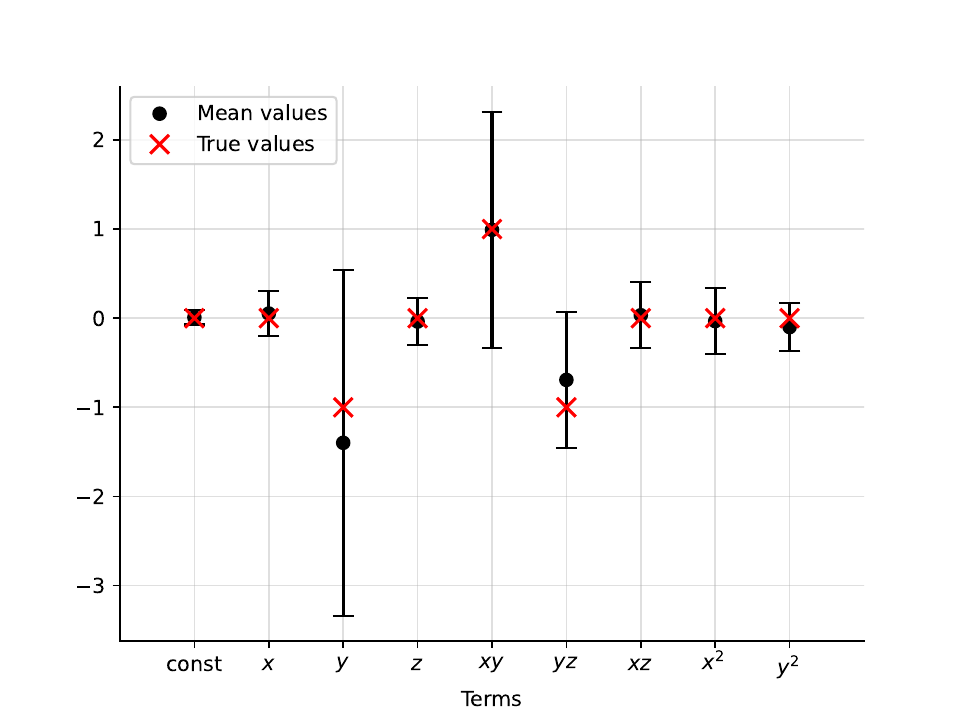}
    \caption{5\% noise \label{fig:LV3DTerms5Noise}}
    \end{subfigure}
    \caption{Means and standard deviations of the learned coefficients for the Lotka-Volterra hybrid neural ODE.}
    \label{fig:LV3DTerms}
\end{figure}

\begin{figure}[H]
    \centering
    \begin{subfigure}{0.49\linewidth}
    \centering
    \includegraphics[width=\linewidth]{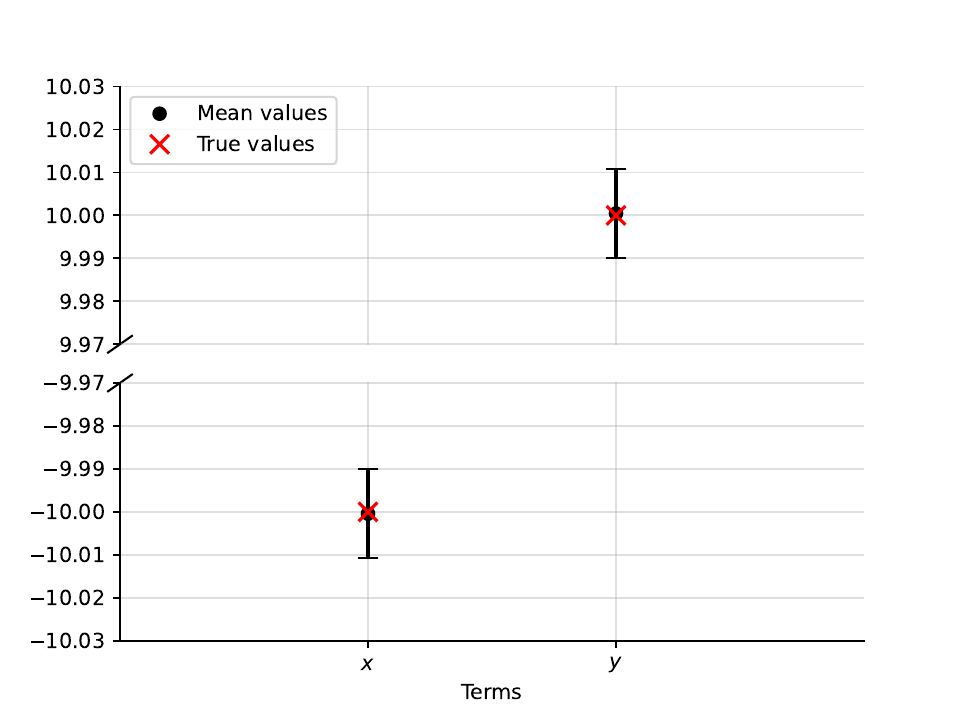}
    \caption{\label{fig:Lorenz5DEq1Terms0Noise} Eq1, 0\% noise}
    \end{subfigure}
    \begin{subfigure}{0.49\linewidth}
    \centering
    \includegraphics[width=\linewidth]{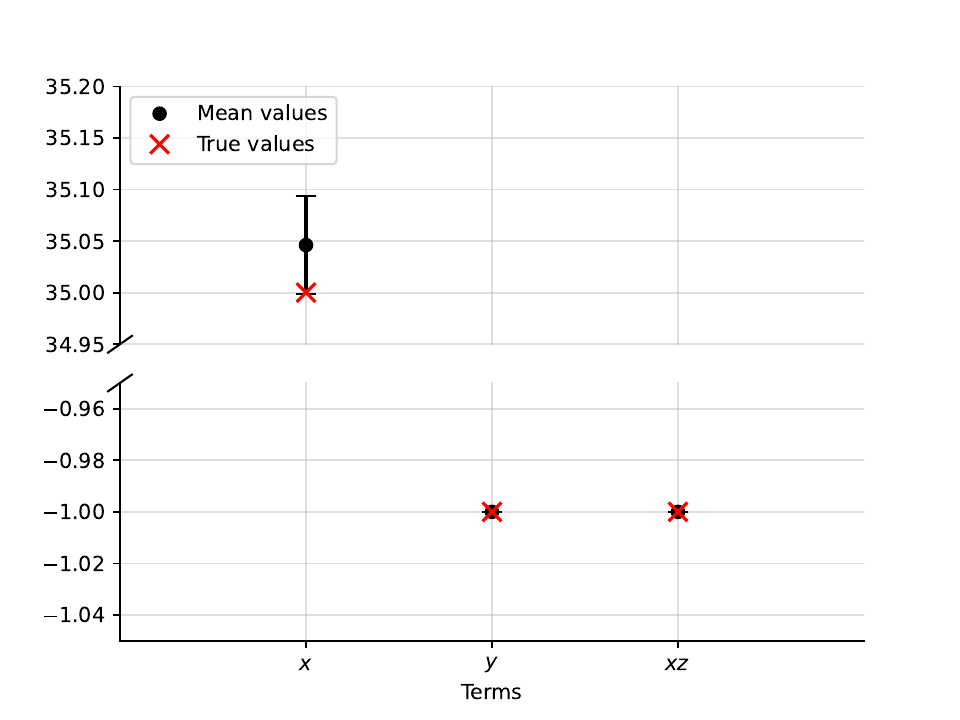}
    \caption{\label{fig:Lorenz5DEq2Terms0Noise} Eq2, 0\% noise}
    \end{subfigure}
    \begin{subfigure}{0.49\linewidth}
    \centering
    \includegraphics[width=\linewidth]{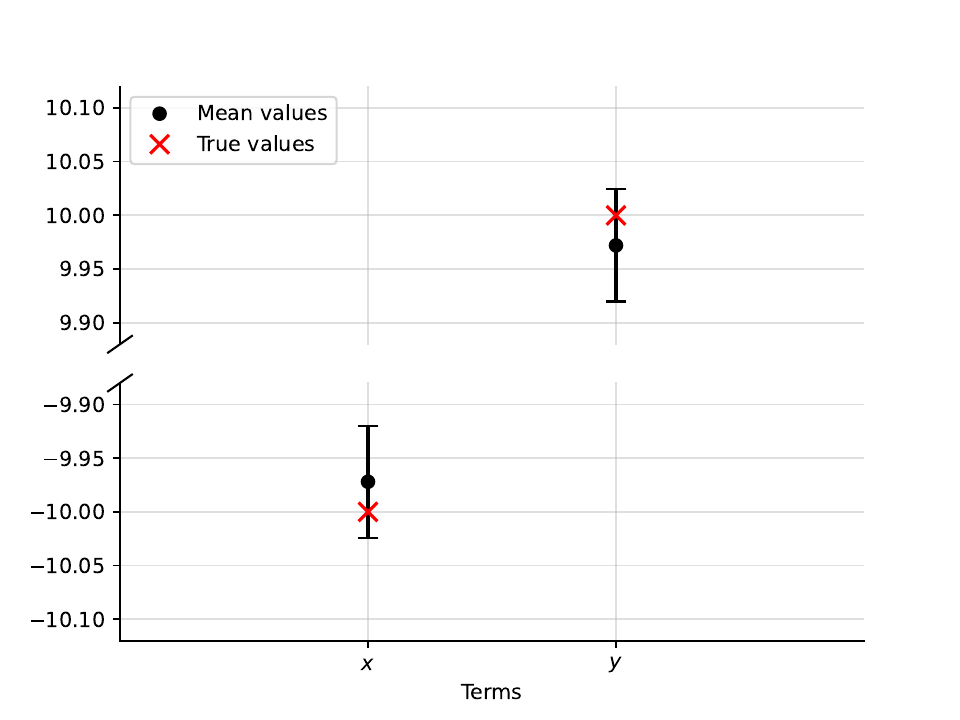}
    \caption{\label{fig:Lorenz5DEq1Terms3Noise} Eq1, 0.3\% noise}
    \end{subfigure}
    \begin{subfigure}{0.49\linewidth}
    \centering
    \includegraphics[width=\linewidth]{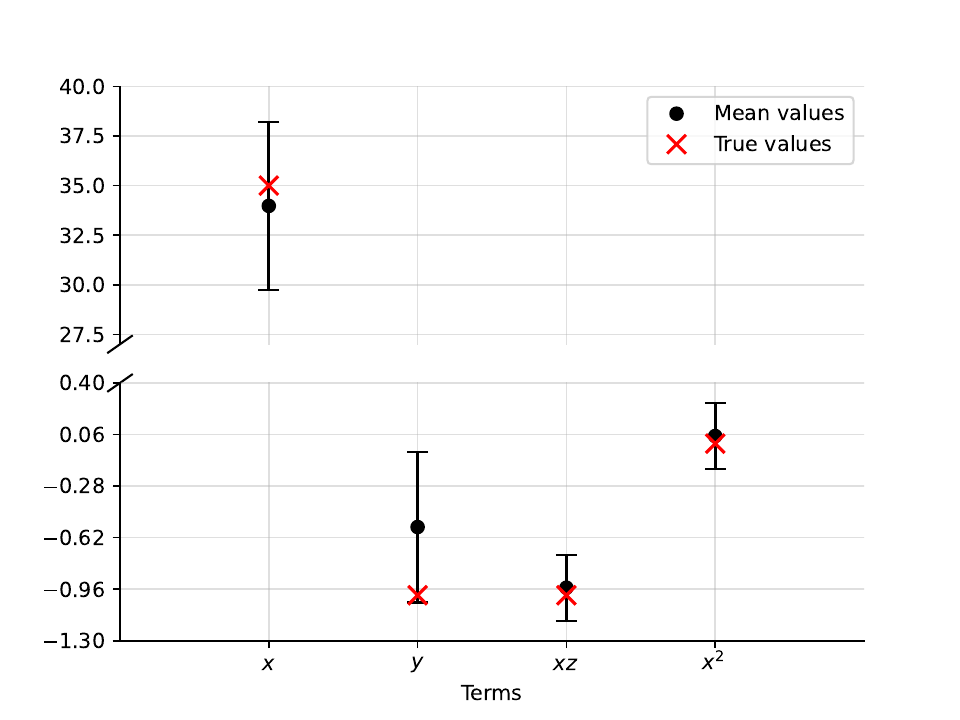}
    \caption{\label{fig:Lorenz5DEq2Terms3Noise} Eq2, 0.3\% noise}
    \end{subfigure}
    \caption{Means and standard deviations of the learned coefficients for the Lorenz hybrid neural ODE.}
    \label{fig:Lorenz5DTerms}
\end{figure}

For this approach, the same case studies are examined, but SR is applied to each model in the ensemble, i.e. prior to averaging. The mean and standard deviations of the coefficients of the terms that appear in any of the learned models are then taken. Here, the ensembles consist of 20 hybrid neural ODEs to obtain more accurate estimates of the means and variances. The results of these analyses are shown in Figures \ref{fig:LV3DTerms} and \ref{fig:Lorenz5DTerms}, while also being summarised in Tables \ref{tab:LV3TermsTable} and \ref{tab:Lorenz5DTermsTable}. In each case, the error bars correspond to one standard deviation.

\begin{table}[H]
\centering
\footnotesize
\begin{tabular}{| c | c | c |}\hline
\textbf{Noise} & \textbf{Mean Learned Equation} & \textbf{True Equation}\\ \hline
0\% & $\dot{y}=-1.00250887y + xy - yz$ & \multirow{3}{2.2cm}{$\dot{y}=-y+xy-yz$}\\
2\% & $\dot{y}=0.0409 - 1.435y + 0.859xy - 0.743yz - 0.056y^2$ &\\
5\% & $\dot{y}=-0.0089 + 0.0517x - 1.398y - 0.037z + 0.991xy - 0.692yz + 0.033xz - 0.033x^2 - 0.1y^2$ &\\
\hline
\end{tabular}
\caption{\label{tab:LV3TermsTable} Lotka-Volterra: mean coefficients of learned equations across each model in the ensemble, at each level of noise.}
\end{table}

Figure \ref{fig:LV3DTerms0Noise} shows that without measurement noise, the correct equation structure for $\frac{dy}{dt}$ is consistently found, while Figures \ref{fig:LV3DTerms2Noise} and \ref{fig:LV3DTerms5Noise} show that increasing measurement noise increases the number of incorrect terms learned by SR. The mean coefficient values of these incorrect terms are, however, close to 0. Apart from the $y$ term in Figure \ref{fig:LV3DTerms0Noise}, all the true values of the coefficients lie within 1 standard deviation of the mean learned values. The means of these learned equations are given in Table \ref{tab:LV3TermsTable}.

\begin{table}[H]
\centering
\footnotesize
\begin{tabular}{| c | c | c |}\hline
\textbf{Noise} & \textbf{Mean Learned Equations} & \textbf{True Equations}\\ \hline
\multirow{2}{0.7cm}{0\%} & $\dot{x}=10.00042325(y-x)$ & \multirow{4}{2.2cm}{$\dot{x}=10(y-x)$\\ $\dot{y}=x(35-z)-y$} \\
& $\dot{y}=x(35.04614365-z)-y$ &\\
\cline{1-2}
\multirow{2}{0.7cm}{0.3\%} & $\dot{x}=9.9719798(y-x)$ &\\
& $\dot{y}=x(33.97495465-0.95z)-0.55y + 0.05x^2$ &\\
\hline
\end{tabular}
\caption{\label{tab:Lorenz5DTermsTable} Lorenz: mean coefficients of learned equations across each model in the ensemble, at each level of noise}
\end{table}

Figures \ref{fig:Lorenz5DEq1Terms0Noise} and \ref{fig:Lorenz5DEq1Terms3Noise} show that the correct form of the $\frac{dx}{dt}$ equation is consistently learned, both with and without noise (and with accurate coefficient values), since no incorrect terms are recovered. Without noise, the $\frac{dy}{dt}$ is also correctly learned consistently (Figure \ref{fig:Lorenz5DEq2Terms0Noise}), whereas in the presence of noise, an incorrect $xz$ term is also recovered. Again, the true coefficient values all lie within 1 standard deviation of the mean learned values. The means of these learned equations are given Table \ref{tab:Lorenz5DTermsTable}.
\bibliographystyle{elsarticle-num-names} 
\bibliography{references}





\end{document}